\documentclass[10pt,twocolumn,letterpaper]{article}
\usepackage[accsupp]{axessibility}  

\usepackage[pagenumbers,algorithms]{wacv} 

\usepackage{fancyhdr}
\fancypagestyle{arxivheader}{
    \fancyhf{} 
    
    \fancyhead[C]{\small Published in the IEEE/CVF Winter Conference on Applications of Computer Vision (WACV) 2026}
}

\definecolor{wacvblue}{rgb}{0.21,0.49,0.74}
\definecolor{lightblue}{RGB}{220,230,241}
\usepackage[pagebackref,breaklinks,colorlinks,allcolors=wacvblue]{hyperref}
\usepackage{graphicx}
\usepackage{amsmath}
\usepackage{amssymb}
\usepackage{booktabs}
\usepackage{algorithm}
\usepackage{algpseudocode}
\usepackage[table]{xcolor}
\usepackage{multirow}
\usepackage{placeins}

\def\wacvPaperID{1253} 
\def\confName{WACV}
\def\confYear{2026}

\title{Unified Control for Inference-Time Guidance of Denoising Diffusion Models}

\author{
Maurya Goyal\textsuperscript{1},
Anuj Singh\textsuperscript{1,2},
Hadi Jamali-Rad\textsuperscript{1,2} \\ \\
\textsuperscript{1}Delft University of Technology, The Netherlands \\
\textsuperscript{2}Shell Global Solutions International B.V., Amsterdam, The Netherlands
}

\begin{document}
\maketitle
\thispagestyle{arxivheader}
\begin{abstract}
   Aligning diffusion model outputs with downstream objectives is essential for improving task-specific performance. Broadly, inference-time training-free approaches for aligning diffusion models can be categorized into two main strategies: \textbf{sampling-based} methods, which explore multiple candidate outputs and select those with higher reward signals, and \textbf{gradient-guided} methods, which use differentiable reward approximations to directly steer the generation process. In this work, we propose a universal algorithm, \textbf{\texttt{UniCoDe}}, which brings together the strengths of sampling and gradient-based guidance into a unified framework. \texttt{UniCoDe} integrates local gradient signals during sampling, thereby addressing the sampling inefficiency inherent in complex reward-based sampling approaches. 
   By cohesively combining these two paradigms, \texttt{UniCoDe} enables more efficient sampling while offering better trade-offs between reward alignment and divergence from the diffusion unconditional prior. Empirical results demonstrate that \texttt{UniCoDe} remains competitive with state-of-the-art baselines across a range of tasks.
   { The code is available at \href{https://github.com/maurya-goyal10/UniCoDe}{https://github.com/maurya-goyal10/UniCoDe}}
\end{abstract}
    
\section{Introduction}
\label{sec:intro}
\begin{figure}[!htbp]
  \centering
  \includegraphics[trim=80 60 0 20, clip, width=1.05\linewidth]{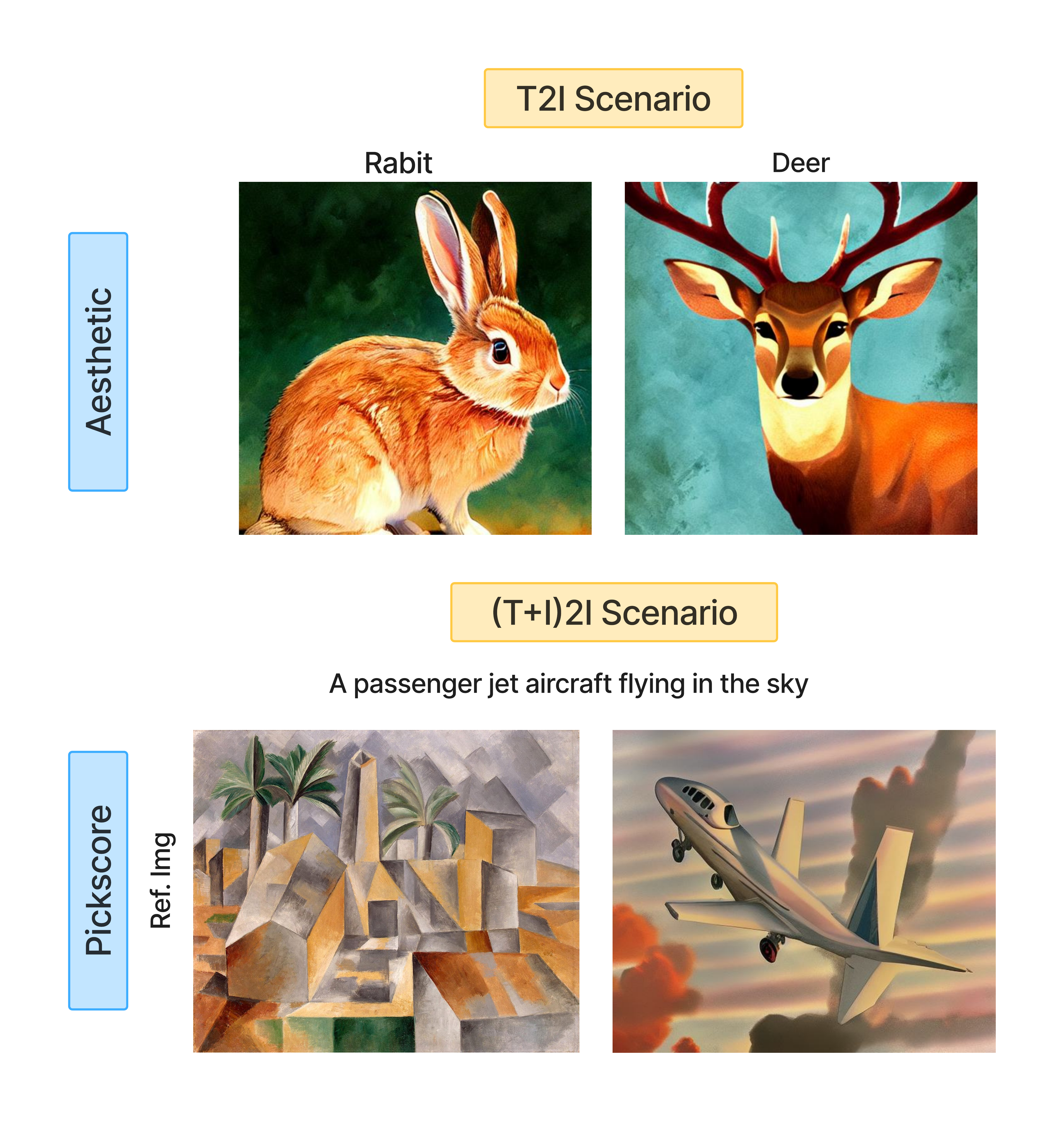}
  \caption{ \texttt{UniCoDe} generates high-quality images across diverse reward objectives such as aesthetic, pickscore, and in (T+I)2I guided using pickscore, demonstrating adaptable performance.}
  \label{fig:teaser}
\end{figure}

Diffusion models \cite{sohl2015deep,ho2020denoising,song2020denoising,song2020score} have illustrated impressive capabilities in sampling complex data distributions, allowing them to generate high-quality images \cite{nichol2021improved,dhariwal2021diffusion, rombach2022high}, videos \cite{bar2024lumiere, oshima2025inference}, audio \cite{kong2020diffwave, huang2023make}, and even biological structures such as proteins and DNA \cite{watson2023novo,jing2022torsional}. These models are typically trained on large and diverse datasets, allowing them to capture a wide range of patterns and variations \cite{blattmann2023stable}. However, in many practical scenarios, it becomes important to \textit{guide} the generative process towards outputs maximizing specific properties, such as aesthetic appeal or adherence with user-provided prompts or images. Thus, conditioning diffusion models to produce outputs that meet user-defined criteria has emerged as a consequential problem. 

There are several ways for conditioning generative models to produce outputs aligned with the user's desired objective. The most straightforward method is fine-tuning \cite{lee2023aligning, prabhudesai2023aligning, black2023training,uehara2024feedback} the unconditional foundational model for each specific task. Fine-tuning has demonstrated strong performance in aligning models with specific tasks, enabling the base model to learn domain-specific nuances, and once fine-tuned, the model can generate aligned outputs rapidly at inference time, requiring no additional computation during sampling. However, fine-tuning is computationally expensive and typically requires substantial labeled data or high-quality reward signals. It also potentially leads to reward over-optimization, a form of reward hacking where the model exploits loopholes in the reward function to produce high-reward outputs that do not align with true user intent. As an alternative, inference-time guidance \cite{song2020score, dhariwal2021diffusion} offers a more flexible solution by enabling training-free conditioning through external reward models during denoising.


In this work, we focus exclusively on inference-time guidance for conditional image generation scenarios, specifically in the setting where the reward model is applied to the final denoised (clean) samples rather than intermediate noisy ones. This allows us to leverage off-the-shelf reward models directly, without modifying them or extra training. The reward models can be any loss function, classifier, or probability estimator that measures how well the output aligns with the given user conditioning signal. 

Training-free guidance methods can be extensively categorized into gradient-based \cite{chung2022diffusion,bansal2023universal, yu2023freedom, he2023manifold, ye2024tfg} and sampling-based approaches \cite{li2024derivative,singh2025code}. Gradient guidance involves perturbing samples during the denoising process in the direction that increases the reward, effectively guiding the generation toward more desirable outcomes. Conversely, sampling-based methods utilize stochasticity by generating multiple candidate samples and preferentially sampling more from those expected to yield higher rewards. While sampling-based approaches often offer a better trade-off between target reward alignment and the divergence from the prior \cite{singh2025code}, they come at a high computational cost, as the number of required samples increases exponentially with the divergence between the prior distribution and the desired posterior distribution \cite{wu2023practical}. On the other hand, gradient-based guidance can directly steer the generation toward desirable outputs but faces limitations in flexibility, differentiability, and generalization. In this work, we explore how these two paradigms can be effectively unified to leverage their complementary strengths and aim to answer the key question: \textbf{Can sampling-based and gradient-based inference-time guidance be unified to reduce sampling overhead, enhance output quality, and achieve better trade-offs between reward alignment and prior divergence? }


Inspired by Controlled Denoising (\texttt{CoDe}) \cite{singh2025code} proposed by Singh \etal, we incorporate blockwise guidance with blockwise sampling to answer this question. This unified framework leverages the strengths of both sampling-based and gradient-based methods: the flexibility of sample-based reward maximization and the efficiency of gradient-driven updates. By using gradient signals to explicitly steer the prior distribution toward the desired posterior, we reduce the number of samples required during the reverse diffusion process, thus improving sampling efficiency while maintaining high output quality and strong alignment with the target conditioning. Furthermore, we also investigate the following set of extensions designed to enhance the overall performance and efficiency of the method, while also improving its generality and adaptability across different tasks and reward structures:

\begin{itemize}
    \item \textbf{Sampling schedule:} Introducing a scheduled sampling strategy that allocates more sampling budget to steps in the denoising process where the reward signal is expected to be most significant.
    \item \textbf{Non-greedy sampling:} Replacing the myopic greedy sampling strategy with more flexible alternatives such as multinomial sampling to better control the exploration-exploitation tradeoff.
    \item \textbf{Clustering-based guidance:} Incorporating clustering techniques to limit the overhead of applying gradient guidance to every sample individually.
    \item \textbf{Non-differentiable rewards:} Also test the method on non-differentiable rewards using zero-order optimization, utilizing the torchopt\cite{torchopt} library.
\end{itemize}

\section{Preliminaries and Related Work}
\label{background}
\textbf{Diffusion Models.} 
Diffusion models are Markov Chains where for the forward process, we iteratively noise a clean input by adding Gaussian noise to it. Thus, as we go from $t{=}0$ (clean image) to $t{=T}$ and for large enough $T$, we reach Gaussian noise. Using a neural network we learn to reverse this process $p_{\theta}(x_{t-1}|x_t)$ (denoising) since $q(x_{t}|x_{t-1})$ is known. Consequently, iteratively applying $p_{\theta}(x_{t-1}|x_t)$ to a randomly sampled Gaussian noise, we can sample from the target distribution or generate a new image. 


\textbf{Score Based Formulation.}
Diffusion models have a score-based formulation through SDE \cite{song2020score} to estimate the score function, which is defined as $\nabla_{x_t} \log p(x_t)$ using a neural network. 
The forward process is defined as:
\begin{equation}
dx = \textbf{f}(\textbf{x},t)dt + g(t)d\mathbf{w}.
  \label{eq:sde-forward)}
\end{equation}
In the above equation $f(x,t)$ is vectored value function called the \textit{drift} coefficient of $x(t)$ and $g(t)$ is a scalar function known as the \textit{diffusion} coefficient of $x(t)$, both of which are known. The reverse process is defined as:
\begin{equation}
dx = [\textbf{f}(\textbf{x},t)dt - g(t)^2\nabla_x \log p_t(\textbf{x})]dt + g(t) d\bar{\mathbf{w}}.
  \label{eq:sde-backward)}
\end{equation}
Where $\bar{\mathbf{w}}$ is the standard Wiener process when time flows backward from $T$ to $0$. The only unknown term is the score function i.e. $\nabla_x \log p_t(x)$ which the neural network learns to predict. Thus after sampling $x(T) \sim p_T$ and reversing the process we can obtain the clean image (sample from $p_0$).

\textbf{Alignment Objective.}
Considering an off-the-shelf reward function \( r: \mathcal{X} \to \mathbb{R} \) that quantifies the quality of alignment of generated samples with a given conditioning. Let \(\pi(\cdot)\) denote the reward-aligned diffusion model and \(p(\cdot)\) the base, unconditional reference diffusion model \cite{singh2025code}. From a reinforcement learning perspective, the alignment objective can be formulated as optimizing \(\pi\) to maximize the expected reward under the aligned distribution relative to \(p\), thereby encouraging the generation of samples satisfying:
\begin{align}
    \pi^{*}_{\lambda} 
    = \arg\max_{\pi} &\Bigg[
        \lambda \, \mathbb{E}_{x_{t-1} \sim \pi(x_{t-1} \mid x_t)} 
        \Big[ \mathbb{E}_{x_0 \sim p(x_{t-1} \mid x_t)} [r(x_0)] \Big] \notag \\
        &\quad - \mathrm{KL} \big( \pi(x_{t-1}|x_t) \, \| \, p(x_{t-1}|x_t) \big)
    \Bigg],
\label{eq:alignment-obj}
\end{align}
where $\lambda \in \mathbb{R}^{\geq0}$ trades off reward for drift from the base diffusion model $p$. 

\textbf{Gradient Based Guidance.} \cite{song2020score, dhariwal2021diffusion}
For conditional generation\cite{dhariwal2021diffusion}, the goal is to sample from $p(x|y)$ instead of $p(x)$. Thus wrt to the score-based formulation we want the $\nabla_{x_t} \log p(x_t|y)$. Using Bayes Rule \cite{luo2022understanding}: $\nabla$ represents $\nabla_{x_t}$
\begin{align}
\nabla \log p(x_t|y) 
&= \nabla \log\left(\frac{p(x_t)p(y|x_t)}{p(y)}\right) \notag \\
&= \nabla \log p(x_t) + \nabla \log p(y|x_t) - \nabla \log p(y) \notag \\
&= \underbrace{\nabla \log p(x_t)}_{\text{unconditional score}} 
+ \underbrace{\nabla \log p(y|x_t)}_{\text{adversarial gradient}}.
\label{eq:gradient_guidance}
\end{align}
In Eq.~\ref{eq:gradient_guidance}, we know the first-term unconditional score from the foundational diffusion model and we get the second term using an off-the-shelf reward model. 



\textbf{Sampling Based Guidance (\texttt{CoDe}).} The idea of using sampling for guidance builds on the Best-of-N (BoN) approach \cite{beirami266741736theoretical}. When performing conditional sampling from the posterior $p(x|y)$ given the prior $p(x)$, we can draw multiple independent samples from the prior and at the end of the denoising select the one that best aligns with the posterior. 
A more aggressive way is to estimate the expected alignment with the posterior
and then explore more in the direction expected to give higher rewards \cite{li2024derivative,singh2025code}. 

Further details on the background and a more comprehensive review of the related literature are provided in the Appendix (refer A.'Related Works' and B.'Background'). 
\section{Methodology}
\begin{figure*}[t]
  \centering
  \includegraphics[trim=0 135 0 130,clip,width=1.0\linewidth]{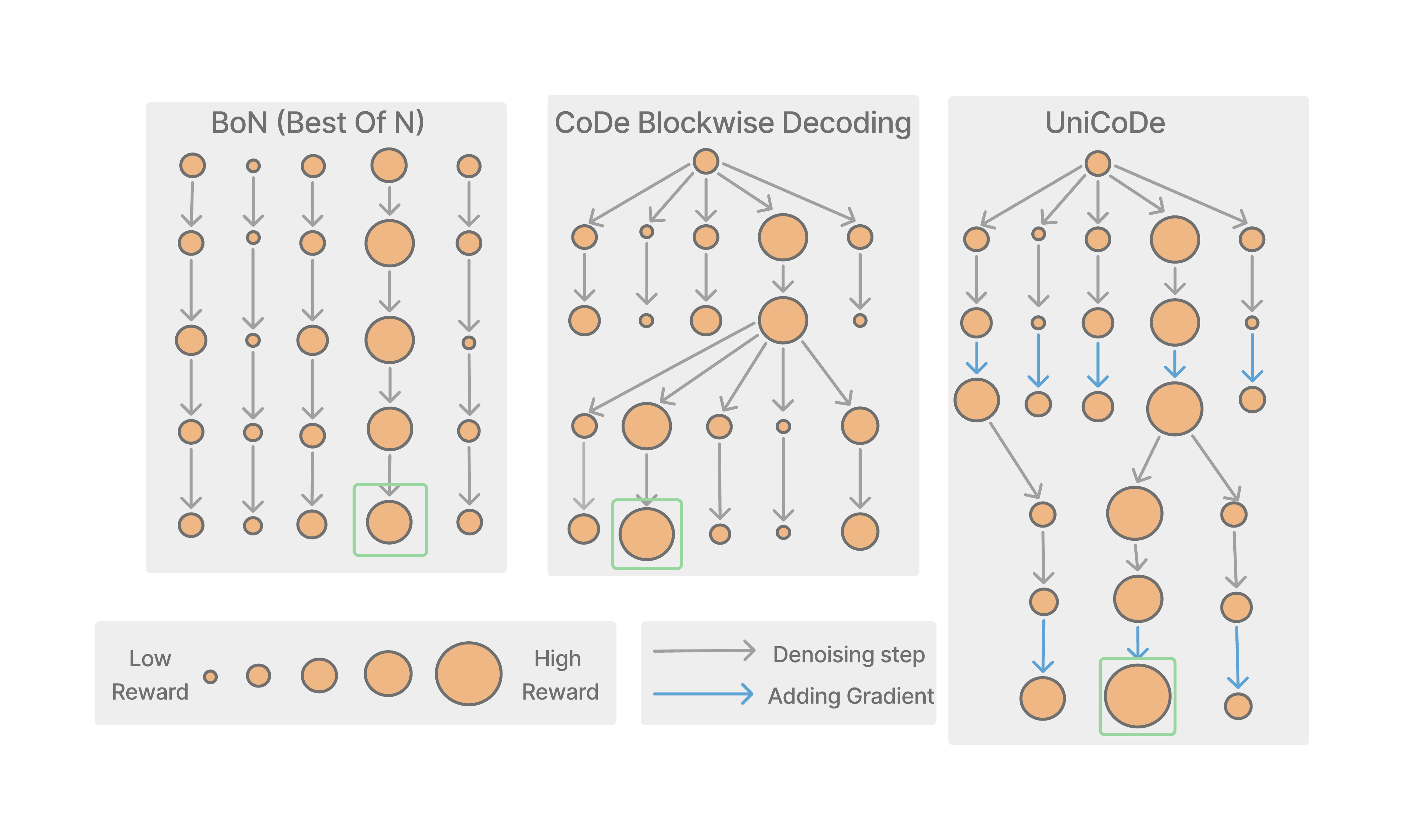}
\caption{ Schematic illustration of sampling-based guidance methods: \textbf{\texttt{BoN}} runs parallel diffusion streams and selects the highest-reward sample after full denoising; \textbf{\texttt{CoDe}} performs greedy blockwise selection to focus on promising regions early; \textbf{\texttt{UniCoDe}} enhances \texttt{CoDe} with blockwise gradient-based guidance, scheduled particle sizes, and multinomial sampling for better efficiency and performance.}
  \label{fig:schematic_diagram}
\end{figure*}

\begin{algorithm}[t]
\caption{\texttt{UniCoDe}}
\label{alg:code+grad}
\begin{algorithmic}[1]
\Require Timesteps: $T$, Samples: $N$, Blocksizes: $B_g$, $B_s$, Temp. $\tau$, $p_{\theta}$, Reward fn. $r$, guidance scale: $\gamma$
\State Sample initial noise: $\{z_T^{(n)}\}_{n=1}^N \sim \mathcal{N}(0, I)$
\State Initialize counter: $s \gets 1$
\For{$t = T, \ldots, 1$}
    \State $\{z_{t-1}^{(n)}\}_{n=1}^N \gets 
    p_{\theta}(\{z_t^{(n)}\}_{n=1}^N, t)$

    \If{$\text{mod}(s, B_g) = 0$} \label{line:5}
        \If{gradient condition satisfied}
            \For{$k = 1$ to $N$} \label{line:7}
                \State $g_k \gets \hyperref[alg:grad]{\texttt{\textsc{Grad}}}(z_{t-1}^{(k)},t-1)$
                \State $z_{t-1}^{(k)} \gets z_{t-1}^{(k)} + \gamma \cdot g_k$
            \EndFor \label{line:10}
        \EndIf
    \EndIf \label{line:12}

    \If{$\text{mod}(s, B_s) = 0$} \label{line:13}
        \State $\{\hat{z}_0^{(n)}\} \gets \mathbb{E}_{p_{\theta}}(\{z_0^{(n)}\} | \{z_{t-1}^{(n)}\},t-1)$ 
        \State $\{z_{t-1}^{(n)}\} \gets \hyperref[alg:sampling]{\texttt{\textsc{Sample}}}(\{z_{t-1}^{(n)}\}, r(D(\{\hat{z}_0^{(n)}\})), \tau)$
    \EndIf \label{line:16}

    \State $s \gets s + 1$
\EndFor
\State \Return $z_0$
\end{algorithmic}
\end{algorithm}

We employ blockwise gradient guidance (Eq.~\ref{eq:gradient_guidance}) to compute and add gradients at the respective timestep for every parallel stream before performing sampling. This setup allows us to maintain multiple candidate trajectories (streams) that explore different directions in the sample space. Among these, the method preferentially explores directions that are expected to yield higher rewards, while the gradient guidance simultaneously nudges each stream closer to the posterior distribution. The methodology is detailed in Algorithm~\ref{alg:code+grad} and illustrated schematically in Fig.~\ref{fig:schematic_diagram}. Specifically, blockwise gradient guidance (Appendix Algorithm 1) is applied independently across streams (lines \ref{line:7}–\ref{line:10}), ensuring localized posterior alignment, after which blockwise sampling (Appendix Algorithm 2) (lines \ref{line:13}–\ref{line:16}) selects and emphasizes high-reward trajectories. In this way, our unified framework (\texttt{UniCoDe}) integrates exploration across multiple streams with posterior alignment via gradient guidance, distinguishing it from prior approaches (\texttt{CoDe},\texttt{BoN}).  

To make the sampling more general we also change the greedy sampling to temperature-based multinomial sampling, this allows the model to consider paths that can give higher rewards later in the denoising and using the temperature we can adjust the aggressiveness of the sampling. Thus it allows the sampling to explore more and can adjust the tradeoff between exploring and exploiting by changing the temperature parameter. 

We also experiment with other extensions like having a sampling schedule that allows exploring more in the denoising steps expected to give more rewards. Moreover, we experiment with clustering to decrease the overhead of adding a gradient to all the sample streams and cluster together points in the latent space that are closer, calculate one gradient for all of them, and add it to all the points in the cluster.

\begin{figure*}
  \centering
  \includegraphics[trim=0 50 0 50,clip,width=1.0\linewidth]{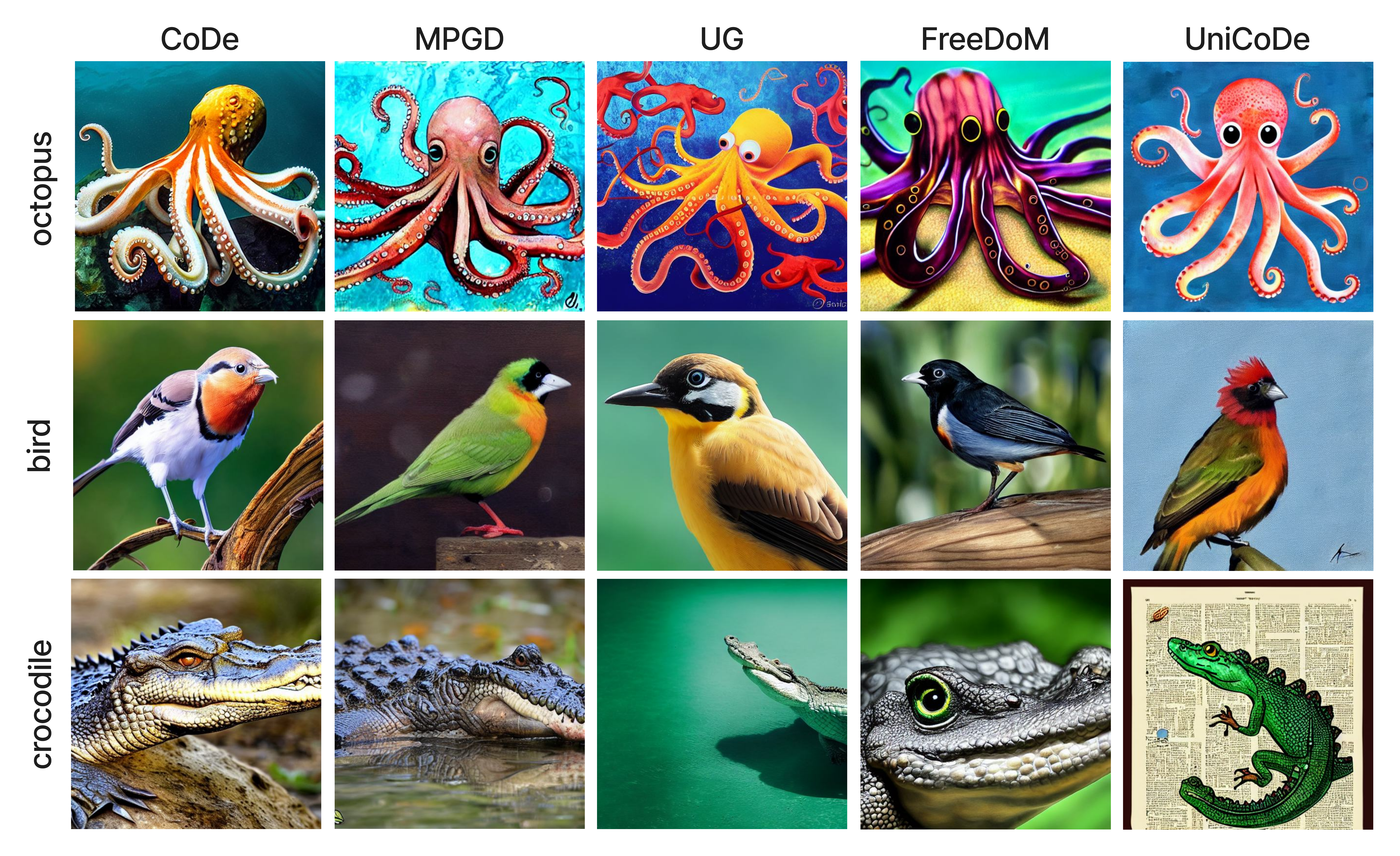}
  \caption{Aesthetic alignment: \texttt{UniCoDe} consistently produces visually pleasing images with strong prompt adherence. While it has higher CMMD (divergence), it avoids reward hacking and tends to yield more aesthetically aligned outputs}
\label{fig:qualitative-aesthetic-final}
\end{figure*}
\section{Experiments}
\label{experiments}


We evaluate the performance of \texttt{UniCoDe} by comparing it against a range of state-of-the-art guidance methods across both Text-to-Image (T2I) and Text-and-Image-to-Image ((T+I)2I) generation tasks. These evaluations are conducted under both differentiable and non-differentiable reward settings to ensure a comprehensive analysis. All experiments are conducted using the pre-trained Stable Diffusion v1.5 model \cite{rombach2022high}, which was trained on the LAION-400M dataset \cite{schuhmann2022laion}. Stable Diffusion is a text-conditioned latent diffusion model trained on paired image-text data. It operates in latent space to perform denoising and leverages text-conditioning to generate prompt aligned images. 

To ensure fair and reproducible comparisons, we generate $10$ images per experimental setting (i.e., for each prompt-reference image pair) using $500$ DDPM steps. All experiments are conducted using NVIDIA A40 GPUs, with the random seed fixed at $2024$ for reproducibility. Both qualitative and quantitative results are reported across a variety of scenarios to highlight the capabilities of \texttt{UniCoDe}. For all scenarios, the sampling block size is set to $5$. The gradient block size is adjusted based on the reward model: $5$ for Aesthetic, $4$ for pickscore, and $2$ for the (T+I)2I setting. Additionally, unless specified otherwise, the guidance scale is fixed at $0.2$ across all experiments.

\textbf{Evaluation Settings and Metrics.}
We evaluate the performance of all methods using a comprehensive set of metrics designed to capture different aspects of alignment, quality, and divergence from the unconditional model. These include the expected reward, Frechet Inception Distance (FID) \cite{heusel2017gans}, CLIP-based Maximum Mean Discrepancy (CMMD) \cite{jayasumana2024rethinking}, I-Gram \cite{gatys2016image, yeh2020improving}, and CLIPScore \cite{hessel2021clipscore}, referred to as T-CLIP throughout the paper. FID and CMMD measure the deviation from the prior distribution defined by the unconditioned Stable Diffusion model, providing a sense of how far the generations diverge from typical outputs. I-Gram captures style and texture similarity between the reference and generated images, particularly relevant in the (T+I)2I setting, while T-CLIP assesses semantic alignment with the input prompt. To support diverse evaluation goals, we consider a variety of reward models: the LAION Aesthetic Predictor v2 \cite{aesthetic, schuhmann2022laion} for image-based aesthetic quality, pickscore \cite{kirstain2023pick} for text-image alignment, a multi-reward configuration that combines aesthetic and pickscore, style transfer, and compressibility \cite{fan2023dpok} a non-differentiable reward facilitating compact image representations. For the quantitative tables, all results are normalized with respect to the base Stable Diffusion model. 

\textbf{Baselines.} In all experimental settings, we compare \texttt{UniCoDe} against \texttt{CoDe} with both $N{=}40$ ($\texttt{CoDe}_{40}$) and $N{=}4$ ($\texttt{CoDe}_{4}$), where $N$ denotes the number of samples generated. We treat the $N{=}4$ configuration as the primary baseline, as \texttt{UniCoDe} itself operates using only $4$ sample streams per block. For the Aesthetic and pickscore reward models, we additionally benchmark against state-of-the-art gradient-based guidance methods, including \texttt{MPGD} \cite{he2023manifold}, \texttt{FreeDoM} \cite{yu2023freedom}, and Universal Guidance (\texttt{UG}) \cite{bansal2023universal}, which are known to improve upon earlier approaches like \texttt{DPS} \cite{chung2022diffusion} by providing more accurate gradient estimates.

\textbf{Notation}
We denote different variants of \texttt{UniCoDe} as follows:
$\texttt{UniCoDe}_{1}$ and $\texttt{UniCoDe}_{2}$ use a fixed sampling size of 4, with greedy and multinomial sampling, respectively.
$\texttt{UniCoDe}_{3}$ and $\texttt{UniCoDe}_{4}$ employ scheduled sampling sizes, again with greedy and multinomial sampling, respectively.
$\texttt{UniCoDe}_{5}$ extends $\texttt{UniCoDe}_{4}$ by incorporating K-Means clustering (K=2) during sampling. Appendix H and I provide hyperparameter analyses that illustrate how these parameters can be effectively controlled. 

\subsection{Image Reward Models}
We first experiment with reward models just based only on the final generated image such as the \textbf{aesthetic scorer}. To guide the diffusion denoising process towards generating aesthetically pleasing images, we use the LAION aesthetic predictor V2 \cite{schuhmann2022laion}, which leverages a multi-layer perceptron (MLP) architecture trained on top of CLIP embeddings. This model’s training data consists of 176,000 human image ratings, spanning a range from 1 to 10, with images achieving a score of 10 being considered art pieces. We evaluate them on the evaluation dataset of animals from ImageNet \cite{deng2009imagenet} (containing 51 prompts). 
 
The quantitative results are summarized in Table \ref{tab:aesthetic_guidance_results}, we can see that \texttt{UniCoDe} not only takes roughly  $4{-}5$ times less time as compared to \texttt{CoDe} \cite{singh2025code} but also gives much better rewards. This comes at the cost of higher divergence from the prior distribution which is shown by the increase in the CMMD, while the FID and the T-CLIP stay similar. We also ablate on the different settings in \texttt{UniCoDe}, as evident from the results the major improvement comes from adding the gradients, and then we can observe little improvements from scheduling samples and making the sampling multinomial. Furthermore, using clustering we can further reduce the time taken at the cost of the rewards.

\begin{table}
\centering
\small
\renewcommand{\arraystretch}{1.0}
\resizebox{\columnwidth}{!}{
\begin{tabular}{|l|c|c|c|c|c|}
\hline
\multirow{2}{*}{\textbf{Method}} & \multicolumn{5}{c|}{\textbf{Aesthetic Guidance}} \\
\cline{2-6}
 & \textbf{Rew. (↑)} & \textbf{FID (↓)} & \textbf{CMMD (↓)} & \textbf{T-CLIP (↑)} & \textbf{Time (↓)} \\
\hline
\texttt{SD1.5} & 1.00 & 1.000 & 1.00 & 1.000 & 1.0 \\
$\texttt{CoDe}_4$ & 1.08 & 1.026 & 1.36 & 1.001 & 21.7 \\
\hline
$\texttt{CoDe}_{40}$ & 1.19 & 1.037 & \textbf{1.92} & 1.003 & 179.8 \\
\rowcolor{lightblue}
$\texttt{UniCoDe}_1$ & 1.60 & 1.059 & 3.24 & 1.003 & 36.5 \\
\rowcolor{lightblue}
$\texttt{UniCoDe}_2$ & 1.60 & 1.048 & 3.28 & 1.003 & 36.0 \\
\rowcolor{lightblue}
$\texttt{UniCoDe}_3$ & 1.61 & 1.053 & 3.17 & 1.002 & 40.4 \\
\rowcolor{lightblue}
$\texttt{UniCoDe}_4$ & \textbf{1.61} & \textbf{1.027} & 3.12 & 1.003 & 41.3 \\
\rowcolor{lightblue}
$\texttt{UniCoDe}_5$ & 1.58 & 1.034 & 2.97 & \textbf{1.005} & \textbf{29.4} \\
\hline
\end{tabular}
}
\caption{Aesthetic guidance comparison, \texttt{UniCoDe} improves aesthetic alignment and reduces runtime $4${–}$5\times$, with higher CMMD but no reward hacking and strong text alignment.}
\label{tab:aesthetic_guidance_results}
\end{table}

We further compare the performance of \texttt{UniCoDe} against state-of-the-art gradient-based guidance methods, as shown in Table~\ref{tab:grad_results_aesthetic}. While \texttt{UniCoDe} achieves higher reward scores, this improvement is accompanied by a greater divergence from the unconditional prior. However, as demonstrated by the qualitative results in Fig.~\ref{fig:qualitative-aesthetic-final}, where \texttt{UniCoDe} is evaluated alongside all competing methods, this increased divergence does not come at the cost of semantic or perceptual alignment. The samples suggest that \texttt{UniCoDe} is capable of generating aesthetically pleasing images that remain faithful to the given prompts, indicating that its strong reward performance is not the result of reward hacking but rather meaningful alignment with the conditioning inputs.

\begin{table}[!htbp]
\centering
\small
\setlength{\tabcolsep}{4pt}
\begin{tabular}{|l|c|c|c|c|}
\hline
\multirow{2}{*}{\textbf{Method}} & \multicolumn{4}{c|}{\textbf{Aesthetic Guidance}} \\
\cline{2-5}
 & \textbf{Rew (↑)} & \textbf{FID (↓)} & \textbf{CMMD (↓)} & \textbf{T-CLIP (↑)} \\
\hline
\texttt{MPGD} & 1.15 & 1.081 & \textbf{2.325} & 0.990\\
\texttt{FreeDoM} & 1.27 &  1.184 & 3.138 & 0.983\\
\texttt{UG} & 1.26 & 1.742 & 3.670 & 0.968  \\
\rowcolor{lightblue}
\texttt{UniCoDe} &\textbf{1.61} & \textbf{1.027} & 3.123 & \textbf{1.003} \\

\hline
\end{tabular}
\caption{Aesthetic guidance results comparing gradient guidance methods to \texttt{UniCoDe}. \texttt{UniCoDe} achieves better reward and better prompt alignment (T-CLIP score)}
\label{tab:grad_results_aesthetic}
\end{table}


\begin{figure*}
  \centering
  \includegraphics[trim=0 90 0 90,clip,width=1.0\linewidth]{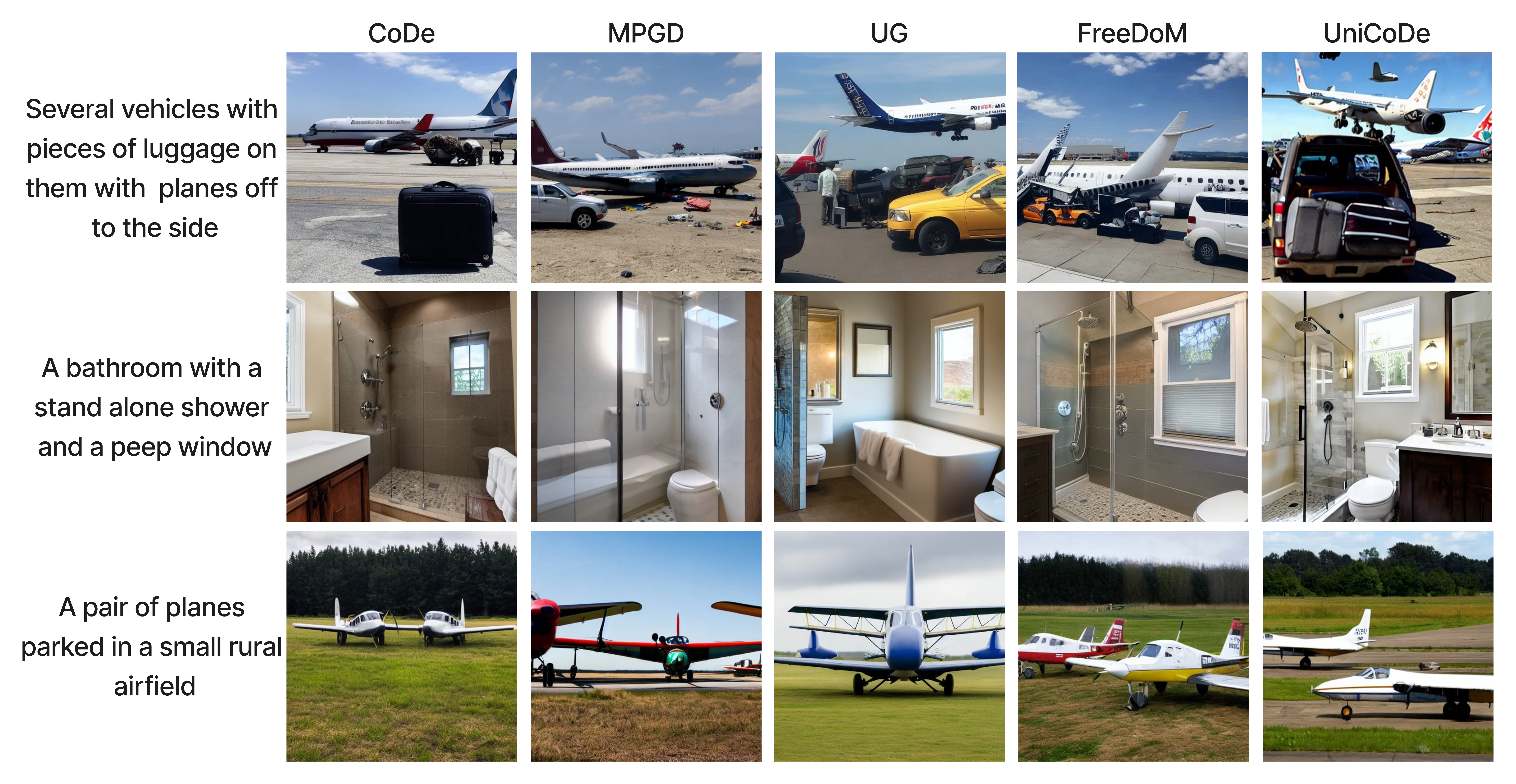}
  \caption{These qualitative examples illustrate how \texttt{UniCoDe} produces images that more accurately and richly reflect the nuances of the input prompts across various scenarios. This is quantitatively supported by higher pickscore and T-CLIP scores (Tables~\ref{tab:grad_results_pickscore}, \ref{tab:pickscore_guidance_results}).}
\label{fig:qualitative-pickscore-1}
\end{figure*}

In this section, we guide the model to adhere better to the given text prompt. Although Stable Diffusion v1.5 is trained on text input conditioning, the performance (adherence to prompt and output quality) starts degrading as the prompt complexity increases. Hence, we use \textbf{pickscore} \cite{kirstain2023pick} as the reward to generate good-quality images adhering to the prompts. Pickscore is a CLIP-based scoring function trained on Pick-a-Pic, a large dataset of text-to-image prompts and real user preferences over generated images. We evaluate the performance of the generated images with prompts taken from the HPDv2 \cite{wu2023human} evaluation settings. 

The quantitative results are given in Tables~\ref{tab:pickscore_guidance_results} and \ref{tab:grad_results_pickscore}, and the qualitative examples are shown in Fig.~\ref{fig:qualitative-pickscore-1}. \texttt{UniCoDe} significantly improves efficiency ($3{-}4\times$ faster), achieves higher rewards, and maintains less divergence from the prior (lower CMMD). Moreover, \texttt{UniCoDe} outperforms gradient guidance methods, providing better prompt alignment and image quality. We also evaluate this setting on a more capable diffusion backbone SD 2.1, and Appendix E Table 1 shows that the same advantages hold.

The tradeoff curves in Fig.~\ref{fig:rew_vs_div} demonstrate that \texttt{UniCoDe} achieves the optimal balance, attaining higher rewards (pickscore) with lower divergence (CMMD) from the prior (unconditional base model) compared to other methods. Further results of this ablation is present in Appendix G.

\begin{table}
\centering
\small
\renewcommand{\arraystretch}{1.0}
\resizebox{\columnwidth}{!}{
\begin{tabular}{|l|c|c|c|c|}
\hline
\multirow{2}{*}{\textbf{Method}} & \multicolumn{4}{c|}{\textbf{Pickscore Guidance (T2I-Based)}} \\
\cline{2-5}
 & \textbf{Rew. (↑)} & \textbf{CMMD (↓)} & \textbf{T-CLIP (↑)} & \textbf{Time (↓)} \\
\hline
\texttt{SD1.5} & 1.000 & 1.000 & 1.000 & 1.0 \\
$\texttt{CoDe}_4$ & 1.054 & 1.220 & 1.008 & 5.7 \\
\hline
$\texttt{CoDe}_{40}$ & 1.107 & 1.611 & 1.016 & 51.2 \\
\rowcolor{lightblue}
$\texttt{UniCoDe}_1$ & 1.116 & 1.469 & \textbf{1.019} & 13.5 \\
\rowcolor{lightblue}
$\texttt{UniCoDe}_2$ & 1.116 & 1.448 & 1.014 & 13.5 \\
\rowcolor{lightblue}
$\texttt{UniCoDe}_3$ & \textbf{1.118} & 1.439 & 1.016 & 13.5 \\
\rowcolor{lightblue}
$\texttt{UniCoDe}_4$ & 1.114 & 1.395 & 1.014 & 13.5 \\
\rowcolor{lightblue}
$\texttt{UniCoDe}_5$ & 1.106 & \textbf{1.334} & 1.013 & \textbf{10.4} \\
\hline
\end{tabular}
}
\caption{T2I-based Pickscore guidance results. \texttt{UniCoDe} improves efficiency (3--4$\times$ faster), prompt alignment (↑ Pickscore, ↑ T-CLIP), and prior preservation (↓ CMMD).}
\label{tab:pickscore_guidance_results}
\end{table}

\begin{table}[!htbp]
\centering
\small
\setlength{\tabcolsep}{4pt}
\begin{tabular}{|l|c|c|c|}
\hline
\multirow{2}{*}{\textbf{Method}} & \multicolumn{3}{c|}{\textbf{Pickscore Guidance}} \\
\cline{2-4}
 & \textbf{Rew (↑)} & \textbf{CMMD (↓)} & \textbf{T-CLIP (↑)} \\
\hline
\texttt{MPGD} & 1.0289 & 1.5747 & 1.0032 \\
\texttt{FreeDoM} & 1.0444 & 1.5207 & 1.0067 \\
\texttt{UG} & 1.0423 & 1.9716 & 1.0056 \\
\rowcolor{lightblue}
\texttt{UniCoDe} & \textbf{1.1183} & \textbf{1.4394} & \textbf{1.0154} \\
\hline
\end{tabular}
\caption{Pickscore guidance results comparing gradient guidance methods to \texttt{UniCoDe}. It deviates less from the prior (lower CMMD) and offers better prompt alignment (higher pickscore and T-CLIP).}
\label{tab:grad_results_pickscore}
\end{table}

\begin{figure}[!htbp]
  \centering
  \includegraphics[trim=0 210 0 210,clip,width=1.0\linewidth]{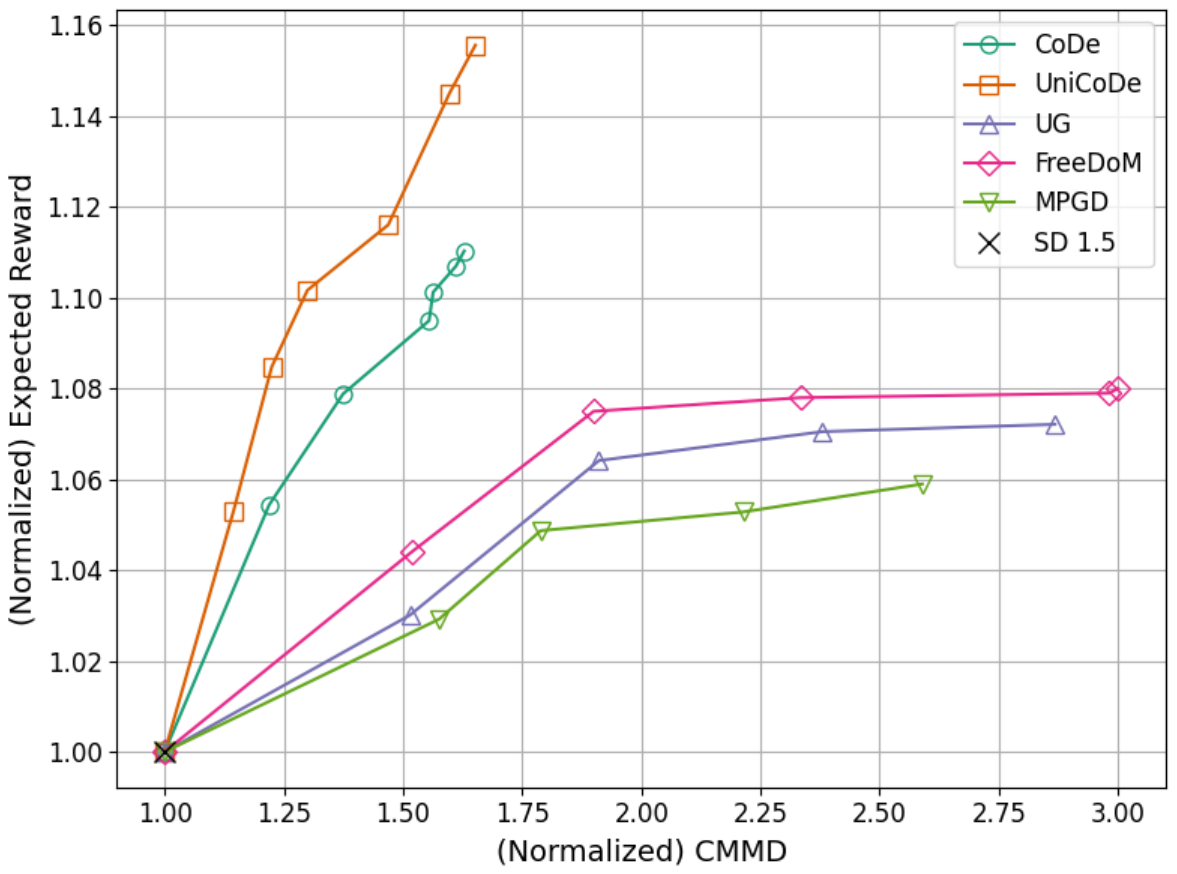}
  \caption{Ablation of rew. (pickscore) vs CMMD (deviation) }
\label{fig:rew_vs_div}
\end{figure}

\subsection{Multi-Reward}
In this section, 
we consider an objective that balances multiple rewards. Specifically, we guide the generation using a weighted sum of the aesthetic score and the pickscore: \textbf{$\boldsymbol{\gamma}_1 \times \text{aesthetic} + \boldsymbol{\gamma}_2 \times \text{pickscore}$}
. The aesthetic score motivates aesthetically pleasing outputs, while the pickscore assures alignment with the prompt based on human evaluations. This grants users a more controlled trade-off between visual quality and prompt fidelity. By adjusting the weights $\gamma_1$ and $\gamma_2$, users can tailor the generation process to their specific preferences.

\begin{figure}[t]
  \centering
  \includegraphics[trim=50 50 0 50,clip,width=1.0\linewidth]{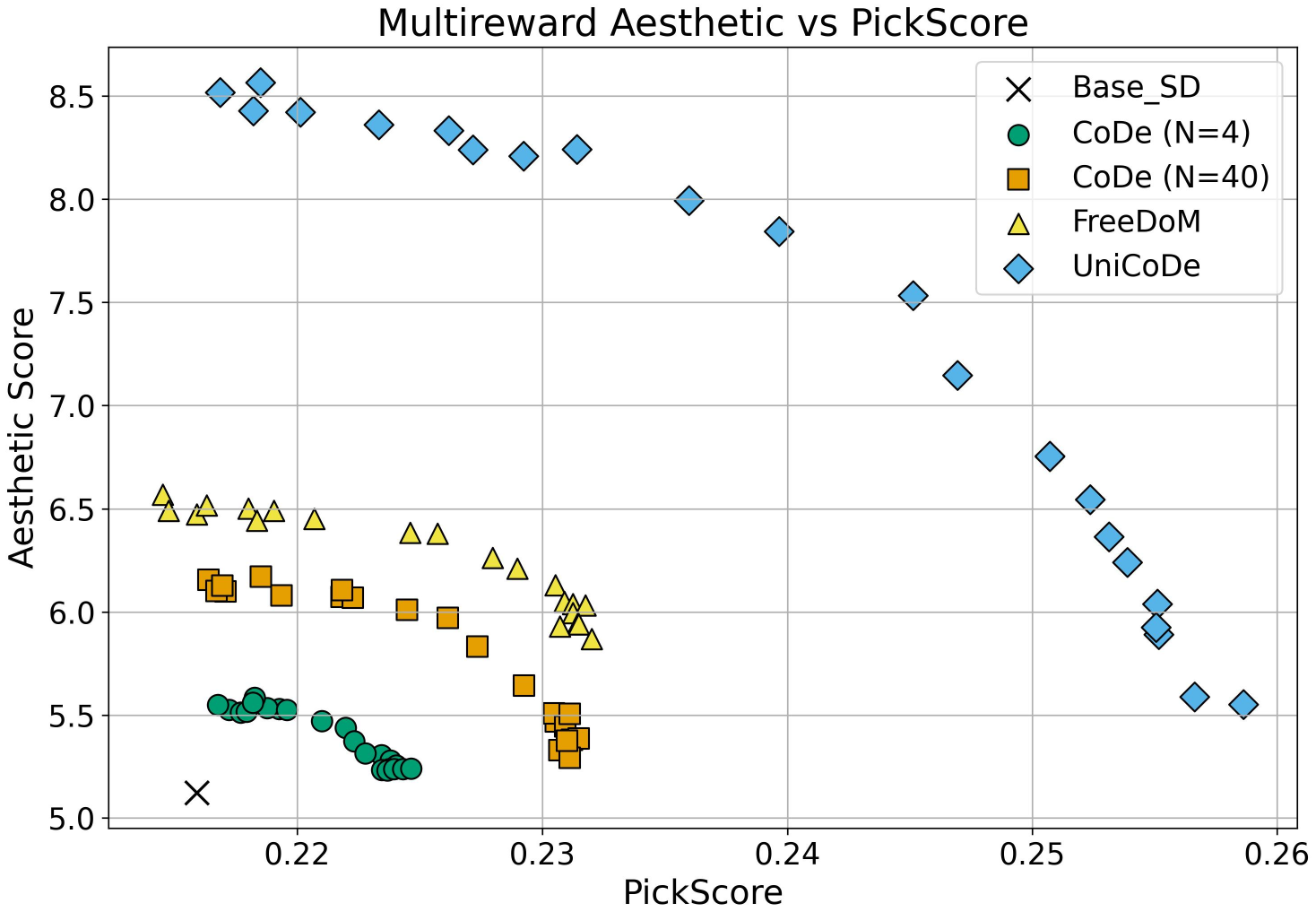}
  \caption{In a multi-reward guidance scenario, \texttt{UniCoDe} consistently achieves higher aesthetic and pickscore, optimizing both rewards compared to other methods. }
\label{fig:multireward}
\end{figure}

A plot of aesthetic score versus pickscore is shown in Fig.~\ref{fig:multireward}. In comparison to the base diffusion model, our approach allows for controlled guidance by adjusting the parameters $\gamma_1$ and $\gamma_2$, enabling a flexible trade-off between multiple reward signals such as aesthetic quality and pickscore. 
The performance obtained by \texttt{UniCoDe} demonstrates significantly improved rewards across multiple reward components. Specifically, \texttt{UniCoDe} significantly improves performance in multi-reward scenarios, boosting the aesthetic reward by about $43\%$ and the pickscore by $7\%$ compared to \texttt{CoDe} on average. Against gradient guidance methods like \texttt{FreeDoM}, \texttt{UniCoDe} still outperforms with increases of $29\%$ (aesthetic) and $6\%$ (pickscore).

\begin{figure*}
  \centering
  \includegraphics[trim=0 210 0 180,clip,width=0.71\linewidth]{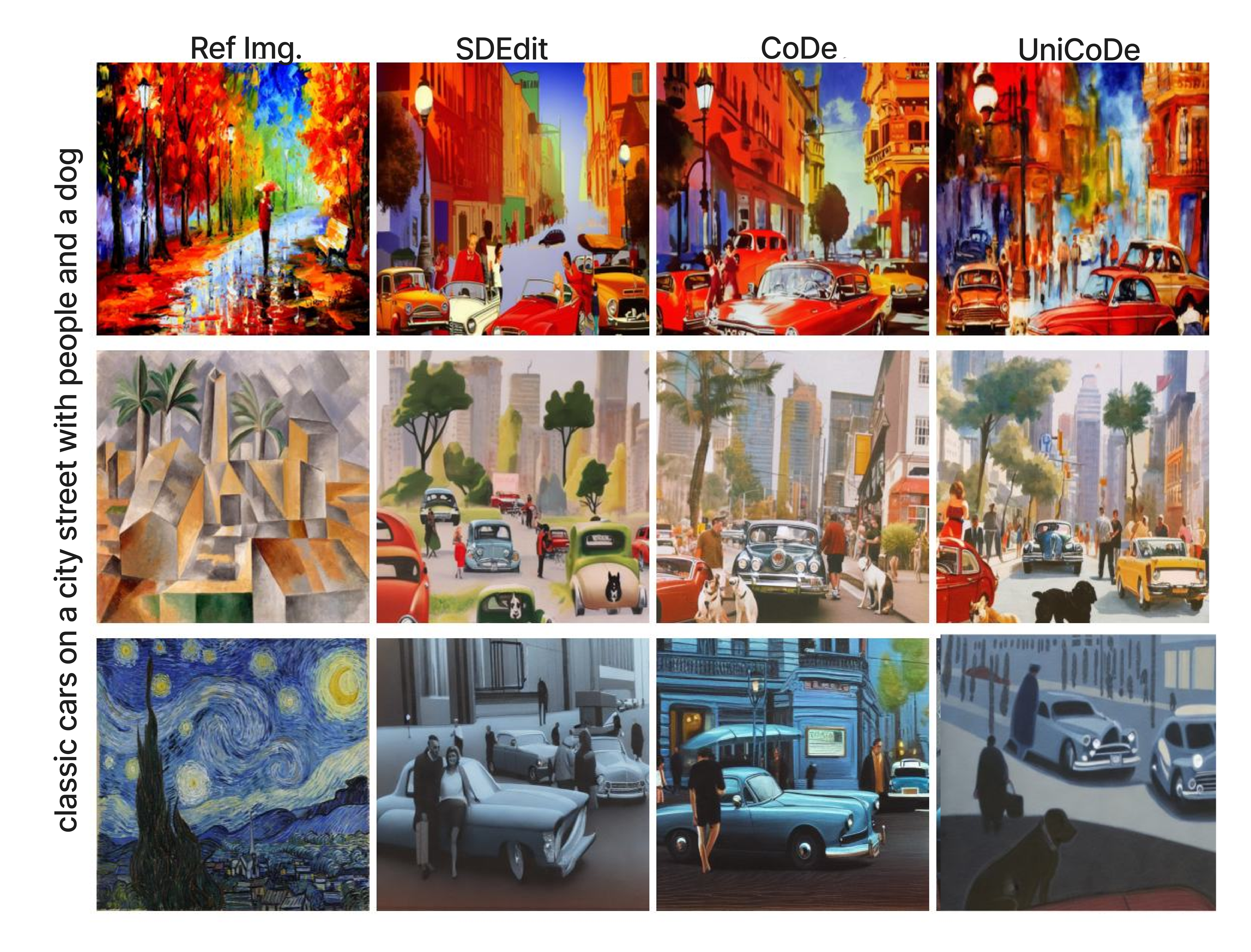}
  \caption{(T+I)2I with pickscore: \texttt{UniCoDe} best balances style and prompt fidelity; \texttt{SDEdit} favors style over prompt adherance}
\label{fig:i2i_qualitative}
\end{figure*}


\subsection{(T+I)2I  (Text-and-Image-to-Image) }
As we move through the experiments we keep on adding more control. This setting blends an additional reference image, enabling the generation of a new image that not only aligns with a given text prompt but also preserves the stylistic elements of the provided reference. We use \texttt{SDEdit} \cite{meng2021sdedit} for the style transfer and start the denoising from an intermediary point (controlled by $\eta$ as $\eta\times T$) by adding noise to the reference image and then during the denoising using the pickscore as the reward to maintain a strong adherence to the prompt.

This offers users more fine-grained control over multiple competing aspects of image generation, specifically balancing adherence to the text prompt with matching the stylistic elements of a reference image. In this case, for prompt adherence, we are guiding the model using the pickscore explicitly while for style transfer it starts the denoising from the noised reference image ($\eta{=}0.6$) \cite{meng2021sdedit}. Thus we want the final generated image to preserve the style and structure of the reference image and at the same time be able to adhere to the provided prompt. As seen in Table \ref{tab:i2i_guidance_results} and Figure~\ref{fig:i2i_qualitative} \texttt{SDEdit} prioritizes style transfer. However due to the absence of guidance its effectiveness in capturing prompt intricacies is limited. In comparison, both \texttt{CoDe} and \texttt{UniCoDe} demonstrate the capacity to balance both style and prompt adherence. \texttt{UniCoDe} achieves a favorable balance between prompt adherence, style transfer, and efficiency, performing slightly better visually while operating at nearly twice the speed of \texttt{CoDe}.

\begin{table}[!htbp]
\centering
\footnotesize
\renewcommand{\arraystretch}{1.1}
\begin{tabular}{|l|c|c|c|c|c|}
\hline
\multirow{2}{*}{\textbf{Method}} & \multicolumn{5}{c|}{\textbf{(T+I)2I Pickscore Guidance}} \\ 
\cline{2-6} & \textbf{Rew.} & \textbf{I-Gr.} & \textbf{CMMD} & \textbf{T-CLIP} & \textbf{Time} \\
\hline
\texttt{SDEdit} & 1.000 & 1.00 & 1.00 & 1.000 & 1.0 \\
$\texttt{CoDe}_4$ & 1.046 & 0.97 & 1.11 & 1.025 & 5.3 \\
\hline
$\texttt{CoDe}_{40}$ & 1.089 & 0.94 & \textbf{1.19} & 1.032 & 47.8 \\
\rowcolor{lightblue}
$\texttt{UniCoDe}_1$ & 1.090 & \textbf{0.97} & 1.24 & \textbf{1.033} & 18.7 \\
\rowcolor{lightblue}
$\texttt{UniCoDe}_2$ & \textbf{1.092} & 0.97 & 1.21 & 1.028 & \textbf{17.5} \\
\hline
\end{tabular}
\caption{\texttt{UniCoDe} improves guidance efficiency (2–3× faster), enhances prompt alignment (Rew., T-CLIP), and better preserves style (I-Gram), with increased prior deviation (CMMD).}
\label{tab:i2i_guidance_results}
\end{table}

\begin{table}[!htbp]
\centering
\footnotesize
\renewcommand{\arraystretch}{1.1}
\begin{tabular}{|l|c|c|c|c|c|}
\hline
\multirow{2}{*}{\textbf{Method}} & \multicolumn{5}{c|}{\textbf{Compression Guidance}} \\ 
\cline{2-6} & \textbf{Rew.} & \textbf{FID} & \textbf{CMMD} & \textbf{T-CLIP} & \textbf{Time} \\
\hline
\texttt{SD 1.5} & 1.000 & 1.000 & 1.000 & 1.000 & 1.00 \\
$\texttt{CoDe}_4$ & 1.676 & 1.132 & 1.806 & 1.004 & 4.99 \\
\hline
$\texttt{CoDe}_{40}$ & 3.219 & \textbf{1.867} & \textbf{5.293} & \textbf{0.982} & \textbf{45.76} \\
\rowcolor{lightblue}
$\texttt{UniCoDe}_1$ \footnotemark & \textbf{3.220} & 2.048 & 5.671 & 0.980 & 238.60 \\
\hline
\end{tabular}
\caption{\texttt{UniCoDe} improves reward and compression performance over \texttt{CoDe}, with slightly reduced T-CLIP alignment and increased compute time.}
\label{tab:compress}
\end{table}



\subsection{Non-Differentiable Rewards}
One of the key advantages of \texttt{CoDe} is its ability to operate in settings with non-differentiable rewards, as it does not rely on gradient-based optimization. Motivated by this, we also extended our algorithm to such scenarios. We use zero-order optimization, which enables gradient approximation without explicitly requiring reward differentiability.

For the non-differentiable reward, we use image compressibility \cite{fan2023dpok} which measures the size of the image in kilobytes. Therefore we guide the model to create lightweight compressible images. The quantitative results are summarized in Table~\ref{tab:compress}. Approximating gradients in this setting is extremely challenging. Even with a large number of forward passes, the estimated gradients remain blurry due to the high dimensionality of the image space. As a result, the efficiency gained by backpropagating through the reward and denoising pipeline is offset by the high computational cost of noisy gradient estimation. This inefficiency is evident from the results in Table~\ref{tab:compress}. In such scenarios, sampling-based guidance methods like \texttt{CoDe} are more practical and effective.
\section{Conclusion and Future Work}
\label{conclusion}

We propose a unified framework that efficiently combines blockwise sampling with blockwise gradient-based guidance, reducing sampling overhead while maintaining strong performance trade-offs. This integration makes \texttt{UniCoDe} more effective for differentiable reward settings, where it outperforms purely sampling-based strategies in both efficiency and reward quality. However, with non-differentiable rewards, gradient estimates become highly noisy, requiring many particles for accuracy, thereby leading to degraded performance compared to purely sampling-based methods in such cases.

Despite progress in conditional image generation, slow inference remains a bottleneck, future work should aim to narrow the gap with unconditional diffusion. Additionally, exploring dynamic temperature scheduling and adaptive particle allocation during sampling may improve generalization and efficiency, potentially boosting reward outcomes across a wider range of tasks. Alternatively, for non-differentiable rewards, training-based gradient approximations like surrogate models could help extend efficiency gains to that domain.



\FloatBarrier
\clearpage
\onecolumn
\appendix

\section*{Supplementary Material}

\section{Related Works}
\label{sec:related-work}

\subsection{Image Generation via Diffusion Models.} Early efforts on image generation relied on GANs \cite{goodfellow2020generative} and VAEs \cite{kingma2013auto,razavi2019generating}. However, these approaches suffer from problems such as training instabilities, limited representational power, and mode collapse. This resulted in the generation of highly smooth images for VAEs \cite{bredell2023explicitly, huang2018introvae, vivekananthan2024comparative} or a lack of sample diversity and control for GANs \cite{srivastava2017veegan, karras2017progressive}. Diffusion models tackle this by gradually transforming noise into clean samples through a learned denoising process iteratively, preserving both the fidelity and diversity of the target distribution \cite{ho2020denoising, dhariwal2021diffusion, rombach2022high}. This makes them particularly well-suited for high-quality and diverse image-generation tasks.

\subsection{Conditional Image Generation.} Image generation models are typically trained on large datasets which enable generating diverse and high-quality images. To provide users with greater control over the generated content, conditional image generation is used. Rather than sampling from the marginal distribution $p(x)$ conditional methods sample from $p(x|c)$, where 
c represents the conditioning signal such as class labels, textual prompts, images, or other forms of guidance. Diffusion models are particularly well-suited for conditional image generation due to their score-based formulation, which estimates the gradient of the data distribution's log density \cite{song2020score}. This formulation naturally accommodates the integration of conditioning signals into the generation process. Conditional image generation can be further classified into two broad categories: training-based and training-free.

\subsection{Training-Based Methods.} This includes methods that train the model with conditioning signals (which can be guided at inference time for better performance using methods like classifier-free guidance (\texttt{CFG})\cite{ho2022classifier}) or fine-tuning-based methods that align the pre-trained foundational model to the specific conditioning. Aligning pre-trained models through fine-tuning is used a lot not just in vision but also for language \cite{ziegler2019fine,ouyang2022training}. For diffusion models, there are several ways of doing that -  direct backpropagation \cite{prabhudesai2023aligning, clark2023directly}, RL-based fine-tuning \cite{fan2023dpok, uehara2024understanding}, preference-based supervised fine-tuning \cite{dang2025personalized, black2023training}, domain adaption \cite{zhang2023adding}, etc. These methods are not scalable and require a lot of computation for fine-tuning each different control signal, therefore, we look into training-free inference-time alignment.

\subsection{Training-Free Methods.} These methods leverage expected rewards during the denoising process to perturb or select optimal samples at \textbf{inference time}. This guidance can be categorized into two main types: gradient guidance and sampling-based guidance.

\subsection{Gradient Guidance.} \cite{chung2022diffusion, song2023loss, bansal2023universal, yu2023freedom, he2023manifold, ye2024tfg} They leverage the expected predicted sample $x_{0|t}$ at each denoising step and compute the gradient of the reward for this estimate. This gradient guides the denoising trajectory toward regions in the sample space that are expected to give higher rewards, effectively \textit{guiding} the generative process to produce samples with improved rewards.

\subsection{Sampling-Based Guidance.} \cite{li2024derivative, singh2025code, kim2025alignment, oshima2025inference} Instead of relying on gradient information, they generate multiple candidate samples at each denoising step and estimate the expected reward for each. This allows them to identify promising directions in the sample space and explore more of those areas in the denoising process, thereby guiding the generation toward samples with higher expected rewards.

\subsection{Combining Gradient and Sampling-Based Guidance.} TDS \cite{wu2023practical} was the first to propose a hybrid approach that combines gradient-based optimization with sampling-based exploration. However, their method relies on the Sequential Monte Carlo (SMC) sampling, assuming the gradient to be sampling from the posterior and is evaluated only on relatively simple tasks such as class-conditional generation on MNIST and CIFAR \cite{krizhevsky2009learning}, limiting its generalizability. In contrast, our work explores a broader range of tasks by leveraging off-the-shelf reward models without restricting the setting to predefined class labels. Concurrent with our work, DAS \cite{kim2025alignment} extends the TDS framework by introducing a tempering scheme to better explore reward-guided generation. However, it does not consider the blockwise perspective in gradient and sampling guidance, nor does it explore complex scenarios such as (T+I)2I (Text-and-Image-
to-Image) guidance.

\subsection{Approximating Gradients for Non-Differential Functions.}
To ensure that the proposed method remains applicable to both differentiable and non-differentiable rewards, gradient approximations are used to enable perturbations in the direction of reward improvement. Two primary strategies exist for this purpose. The first involves training a surrogate model to approximate the reward function, however, this approach conflicts with the overarching goal of preserving a training-free framework. Therefore, this work adopts the second strategy: zero-order optimization \cite{chen2017zoo, liu2020primer}, which allows for gradient approximation using only forward evaluations of the reward function, thereby avoiding the need for additional model training.

\section{Additional Background}
\label{background}

\subsection{Noise Based Formulation.}
The forward process at each time is defined in Eq.~\ref{eq:q(t|t-1)} where the $\beta_t$ represents the variance schedule. ${\alpha}_t = 1 - {\beta}_t$ and $\bar{\alpha}_t = \prod_{t=1}^T{\alpha}_t$ \cite{ho2020denoising}
\begin{equation}
  q(x_t|x_{t-1}) = \mathcal{N}(x_t;\sqrt{1-{\beta}_t}x_{t-1},\beta_t \mathbf{I})
  \label{eq:q(t|t-1)}
\end{equation}
\begin{equation}
  x_t = \sqrt{\bar{\alpha}_t}x_0 + \sqrt{1-\bar{\alpha}_t}\epsilon_t
  \label{eq:x_t|x_0}
\end{equation}
The whole forward process boils down to: 
\begin{equation}
  q(x_{1:T}|x_{t-1}) = \prod_{t=1}^T q(x_{t}|x_{t-1}).
  \label{eq:forward_process)}
\end{equation}
The diffusion model learns the reverse of this forward step:
\begin{equation}
p_{\theta}(x_{t-1}|x_t) = \mathcal{N}(x_{t-1};\mu_{\theta}(x_t,t),{\beta}_t \mathbf{I}),
  \label{eq:p_theta)}
\end{equation}
where the $\mu_{\theta}(x_t,t)$ is defined as: 
\begin{equation}
\mu_{\theta}(x_t,t) = \frac{1}{\sqrt{{\alpha}_t}}\left(x_t - \frac{1-{\alpha}_t}{\sqrt{1-\bar{\alpha}_t}}{\epsilon}_{\theta}(x_t,t)\right).
  \label{eq:mu_theta)}
\end{equation}
Hence, the UNet \cite{ronneberger2015u} given the noisy input ($x_t$) predicts the noise to be removed to obtain the clean sample ($x_0$) from Eq.~\ref{eq:x_t|x_0} as:
\begin{equation}
\epsilon_{\theta}(x_t,t) \approx \epsilon_t = \frac{x_t - \sqrt{\bar{\alpha}_t}x_0}{\sqrt{1-\bar{\alpha}_t}}.
  \label{eq:epsilon)}
\end{equation}

\subsection{Tweedies Formula.} For classifier-based guidance, we need the value of $p(y|x_t)$, which represents the probability of the desired outcome given a noisy image $x_t$. However, most reward functions available off the shelf are not designed to handle noisy inputs, rather they expect clean images. To address this issue, there are two possible approaches: either train a new reward function capable of operating directly on noisy images or estimate the clean image from the noisy input using Tweedie’s formula.\cite{efron2011tweedie}. Since our goal is to develop a training-free method, we choose the second approach. Tweedie’s formula lets us estimate the expected clean image given the noisy input and is given as:

\begin{equation}
\hat{x}_0 = \mathbb{E}[x_0|x_t] = \frac{x_t - \sqrt{1-\bar{\alpha}_t}\epsilon_{\theta}(x_t,t)}{\sqrt{\bar{\alpha}_t}}.
\label{eq:tweedies}
\end{equation}
Using Eq.~\ref{eq:tweedies} we get the reward value as:
\begin{equation}
r(x_{0|t},y) \approx r(\mathbb{E}[x_0|x_t],y) = r(\hat{x}_0,y) \label{eq:reward_tweedies}.
\end{equation}

\subsection{Controlled Decoding (\texttt{CoDe}).} 
To enhance computational efficiency and scalability, \texttt{CoDe} proposes performing this sampling procedure in a blockwise manner, executing it every $B$ step and generating $N$ candidate samples per block. This parallels RL-based objectives where one sample multiple times from a base policy (the base unconditioned diffusion model) and selects the sample best aligned with the reward function.

\subsection{SDEdit.} Stochastic Differential Editing  \cite{meng2021sdedit} is an image synthesis and editing method that generates images aligning with a reference without relying on complex reward models. SDEdit modifies the standard denoising diffusion process by replacing the typical starting point of denoising, $T$, with $r \times T$, where $r$ is a user-defined percentage of noise $r\in(0,1)$. This implies that instead of beginning the denoising from pure random noise, the process initiates from an intermediate point. This is achieved by adding a controlled amount of noise to the reference image, effectively mimicking the forward diffusion process. The choice of $r$ directly influences the output: a higher $r$ allows for greater creative freedom, while a lower $r$ enables the preservation of more structural and stylistic properties from the reference image. SDEdit is employed for (T+I)2I (Text-and-Image-to-Image) generation, enabling style transfer while offering users multiple adjustable parameters to control different aspects of the output image.

\subsection{Zero-Order Optimization.}
Zero-order optimization is used when we have to approximate the $\nabla_x f(x)$ using just the forward pass $f(x)$. 
\begin{equation}
\nabla f(x) \approx \frac{1}{N^{\prime}} \sum_{i=1}^{N^{\prime}} \frac{f(x+\sigma \epsilon_i) - f(x - \sigma \epsilon_i)}{2 \sigma} \epsilon_i,
\label{eq:zoo}
\end{equation}
where $\epsilon$ is a n-dimensional vector for n-dimensional samples $x$ sampled from $(0,\mathbf{I}_d)$ and $\sigma$ is a scalar constant. $N^{\prime}$ is the number of samples the more samples we have the better the estimate at the cost of time and computation. 

\section{Guidance Rescaling}
We rescale the guidance scale using the same mechanism as the one used in \texttt{FreeDoM} \cite{yu2023freedom}. What they do is that they basically guide the model guidance scale times in the direction of the gradient and rescale this scale based on the CFG guidance. Thus they guide the model even more if there is a big difference between the text conditional and the unconditional noise prediction. 

\begin{equation}
\label{eq:scaled_guidance}
\text{scale}_{\text{new}} = \frac{\parallel \text{correction} \parallel_2 \cdot \text{scale}_{\text{CFG}} \cdot \text{scale}_{\text{grad}}}{ \parallel\text{grad} \parallel_2 + \varepsilon}
\end{equation}

where correction is just the CFG \cite{ho2022classifier} correction term:
\begin{equation}
\label{eq:correction}
\text{correction} = \hat{\epsilon}_{\theta}(x_t,prompt) - \hat{\epsilon}_{\theta}(x_t) 
\end{equation}

We found the dynamic rescaling strategy to be the most effective within our evaluation setting and thus adopted it in place of a fixed guidance scale. By normalizing the guidance using the gradient norm and scaling it proportionally to the correction norm it reduces the sensitivity to the choice of guidance scale. Therefore, a consistent range of values (generally between $0.2$ and $0.6$) performs reliably well across different reward models. This not only improved performance stability but also saved considerable time, as it decreased the need to manually tune the guidance scale for each individual reward function checking for a ton of different values based on the reward scale.



\section{Algorithms}
    The algorithms for sampling and gradient based guidance used are as follows:
    
    \begin{algorithm}[H]
    \caption{\texttt{\textsc{Grad}}$(z_t, t)$}
    \label{alg:grad}
    \begin{algorithmic}[1]
    \Require latent $z$, timestep $t$
    \State $\hat{x}_0 \gets D(\mathbb{E}_{p_{\theta}}(z_0 | z_{t},t))$ \Comment{ Expected clean sample}
    \State $\hat{r}(z_t) \gets r(\hat{x}_0)$ \Comment{Compute reward}
    \State $g \gets \nabla_{z_t} \, \hat{r}(z_t)$ \Comment{Compute gradient w.r.t. $z_t$}
    \State \Return $g$
    \end{algorithmic}
    \end{algorithm}
    
    \begin{algorithm}[H]
    \caption{\texttt{\textsc{Sample}}$(\{z_{t-1}^{(n)}\}, r(D(\{\hat{z}_0^{(n)}\})), \tau)$}
    \label{alg:sampling}
    \begin{algorithmic}[1]
    \Require current images $\{z_{t-1}^{(n)}\}_{n=1}^N$, reward vector $\mathbf{r} = r(D(\{\hat{z}_0^{(n)}\}))$, temperature $\tau$, empty list $\{z_{temp}^{(n)}\}_{n=1}^N$
    \State Compute softmax probabilities:
    \[
    P_n \gets \frac{\exp(\mathbf{r}_n / \tau)}{\sum_{j=1}^N \exp(\mathbf{r}_j / \tau)} \quad \text{for } n = 1, \dots, N
    \]
    \For{$i = 1$ to $N$}
        \State Sample index $i \sim \text{Multinomial}(\{P_n\}_{n=1}^N)$
        \State Append $z_{t-1}^{(i)}$ to $\{z_{temp}^{(n)}\}$
    \EndFor
    \State \Return $\{z_{temp}^{(n)}\}_{n=1}^N$
    \end{algorithmic}
    \end{algorithm}

\section{Robustness of \texttt{UniCoDe} Across More Capable Diffusion Models (SD2.1)}
We further evaluate whether the efficiency and alignment benefits of \texttt{UniCoDe} hold when applied to a stronger base model, namely Stable Diffusion $2.1$. As shown in Table \ref{tab:pickscore_results}, the trends observed with weaker diffusion priors consistently translate to this upgraded setting. While \texttt{CoDe}
 achieves a slightly higher pickscore reward, \texttt{UniCoDe} requires nearly $4\times$ less sampling time yet remains highly competitive in prompt adherence. More importantly, \texttt{UniCoDe} exhibits lower CMMD, this demonstrates that \texttt{UniCoDe} maintains its core advantages of efficiency, improved prior preservation, and strong prompt alignment even when operating on higher-capacity diffusion models such as \texttt{SD 2.1}.
\begin{table}[!htbp]
\centering
\footnotesize
\renewcommand{\arraystretch}{1.1}
\begin{tabular}{|l|c|c|c|c|}
\hline
\multirow{2}{*}{\textbf{Method}} & \multicolumn{4}{c|}{\textbf{Prompt Alignment (T2I)}} \\ 
\cline{2-5} & \textbf{Pickscore} & \textbf{CMMD} & \textbf{CLIP} & \textbf{Time} \\
\hline
\texttt{SD2.1} & 1.000 & 1.000 & 1.00 & 1.00 \\
$\texttt{CoDe}_{30}$ & \textbf{1.092} & 4.415 & \textbf{1.012} & 43.39 \\
\rowcolor{lightblue}
$\texttt{UniCoDe}_{4}$ & 1.091 & \textbf{3.961} & 1.010 & \textbf{11.88} \\
\hline
\end{tabular}
\caption{Results on Stable Diffusion 2.1 under prompt alignment (T2I) guidance. \texttt{UniCoDe} provides comparable prompt adherence with substantially lower computation time and reduced prior deviation compared to \texttt{CoDe}.}
\label{tab:pickscore_results}
\end{table}

\section{Hyperparameter Setting}
\subsection{Image Based Guidance}
For the aesthetic reward model, we use the ViT-L/14 as the backbone model for the CLIP. We use a blocksize of $5$ for both the sampling and gradient guidance in this case and set the guidance scale to be $0.2$. We do the denoising for $500$ DDPM steps and the CFG guidance scale is also set to $5$. Also, we start adding the gradients from the $0.6$ noise ratio i.e. if $T=1000$ we start adding the gradients from the $600$ timestep. We use the schedule $[2, 2, 2, 4, 4, 4, 4, 6, 6, 6]$, which allocates a larger sampling budget toward the later stages of denoising. This design aligns with the intuition that the denoising process progressively refines the sample, moving from coarse to fine details. We selected this particular schedule after ablation studies on several alternatives, as it consistently offered the best performance. This preserves the image structure, doesn't reward hack, and guides it towards achieving a better reward. The evaluation set contains $51$ prompts (from the ImageNet evaluation set) and we generate $10$ images for each prompt.

For gradient guidance experiments for \texttt{MPGD} and \texttt{FreeDoM} we do the guidance only for noise ratios between $0.7$ and $0.3$. Thus for timesteps $70{-}30$ in a $100$ step DDIM scheduler. The guidance scales used are $7.5$ for \texttt{MPGD} and $0.2$ for \texttt{FreeDoM}. Additionally, for \texttt{FreeDoM}, we perform $10$ optimization iterations per denoising step. For the \texttt{UG} baseline, we also use $100$ DDIM steps, with $6$ optimization steps and a forward guidance weight of $30$.

\subsection{T2I Setting}
For the pickscore reward model, we again use the ViT-L/14 as the backbone model for the CLIP. We use a blocksize of $5$ for both the sampling and $4$ gradient guidance in this case and set the guidance scale to be $0.2$. We do the denoising for $500$ DDPM steps and the CFG guidance scale is also set to $5$. We do the gradient addition during the whole denoising as we also want to alter the structural properties for the alignment to complex prompts. This preserves the image structure, doesn't reward hack, and guides it towards achieving a better reward. The schedule used is $[2, 6, 6, 2, 2, 2, 4, 4, 6, 6]$ as it preserves coarser details like the spatial structure early on and still preserves the quality. The evaluation set contains $50$ prompts (from the \texttt{HPD} \cite{wu2023human} evaluation dataset) and we generate $10$ images for each prompt. 

The setting for gradient guidance remains the same except for \texttt{UG} where we increase the weight to $150$.  

\subsection{Multireward Setting}
In the multireward setting, we use the same prompts as the Aesthetic case but only a subset of $6$ out of the $51$ and generate $10$ images for each as it takes longer due to the weighted addition of the two reward models. The rest of the settings are the same and we add the gradients during the whole denoising.The values of $\gamma_1$ and $\gamma_2$ are taken as $(1,0), (0,1)$ and we set $\gamma_1$ as $1$ and change $\gamma_2$ in $[2, 3, 5, 10, 15, 20, 25, 30, 50, 70, 100, 150, 200, 250, 300, 350, 400, 450, 500, 750, 1000]$

\subsection{(T+I)2I  Setting}
For this, we use the same $50$ prompts as the T2I setting, use the three style images as the reference images, and generate $10$ images for each prompt and each style. We use the $100$ step DDPM scheduler and we set $\eta$ to be $0.6$. The sampling blocksize is $5$ and the gradient blocksize is $2$ with a guidance scale of $0.4$. The rest of the settings are the same. 

\subsection{Non-Differentiable Reward (Compressibility)}
For this experiment, we use a guidance scale of $0.2$ and perform an ablation over the number of forward passes, ranging from 1 to 50. We observe that increasing the number of forward passes improves gradient stability; however, even at 50 passes, the gradient remains noisy and computational cost increases significantly. To address this, we set the number of sampling streams to $N{=}35$, allowing performance gains to arise more from exploring diverse directions rather than relying solely on gradient information. Evaluation is conducted using four simple prompts: "monkey", "llama", "wolf", and "butterfly", generating 10 samples for each case. 

\section{Ablation Tradeoff Curves for T2I Scenario}
\begin{figure}[ht]
  \centering
  \begin{subfigure}[b]{0.48\linewidth}
    \includegraphics[trim=10 200 50 200 ,width=1.0\linewidth]{rew_vs_div.pdf}
    \caption{Trade off curves for Reward vs Divergence}
      \label{fig:rew_vs_div}
  \end{subfigure}
  \hfill
  \begin{subfigure}[b]{0.48\linewidth}
    \includegraphics[trim=10 200 50 200 ,width=1.0\linewidth]{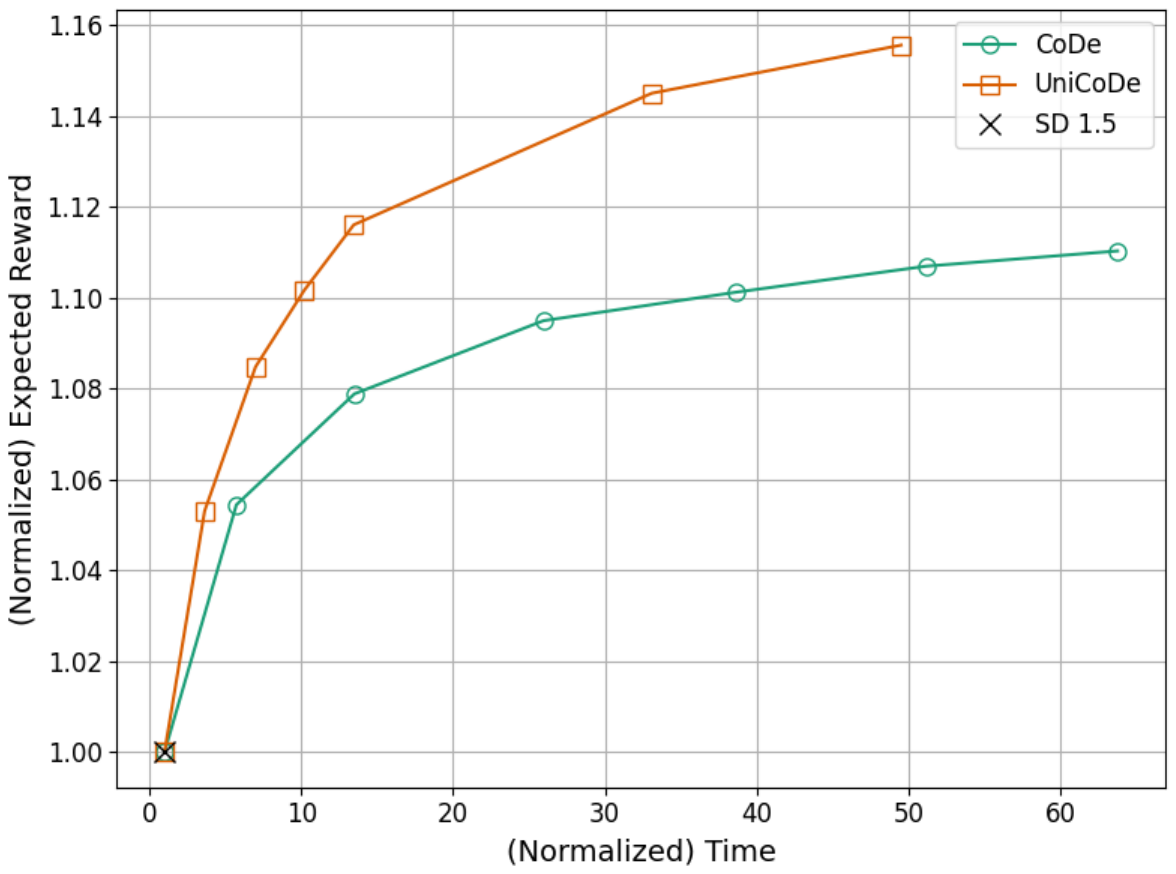}
    \caption{Trade off curves for Reward vs Time}
  \label{fig:rew_vs_time}
  \end{subfigure}
  \label{fig:ablations}
\end{figure}

For the text-to-image (T2I) guidance scenario using pickscore as the reward, we present ablation trade-off curves by varying the number of samples $N$ for both \texttt{CoDe} and \texttt{UniCoDe}. In the gradient-based guidance scenario, we vary either the guidance scale or the number of recurrent time steps. These experiments are designed to evaluate how the proposed method performs across the overall trade-off frontier. We compare performance across multiple axes, including reward versus divergence, reward versus compute, divergence versus compute, and T-CLIP score versus divergence. The corresponding results are presented in Figures~\ref{fig:rew_vs_div}, \ref{fig:rew_vs_time}. 

\section{Blocksize Hyperparameter Analysis}
For all these experiments we use the pickscore reward function in the T2I scenario with a fixed number of samples ($N={4}$). 

\subsection{Equal Blocksizes}
\begin{figure}[ht]
  \centering
  \begin{subfigure}[b]{0.48\linewidth}
    \includegraphics[trim=100 250 100 250 ,width=1.0\linewidth]{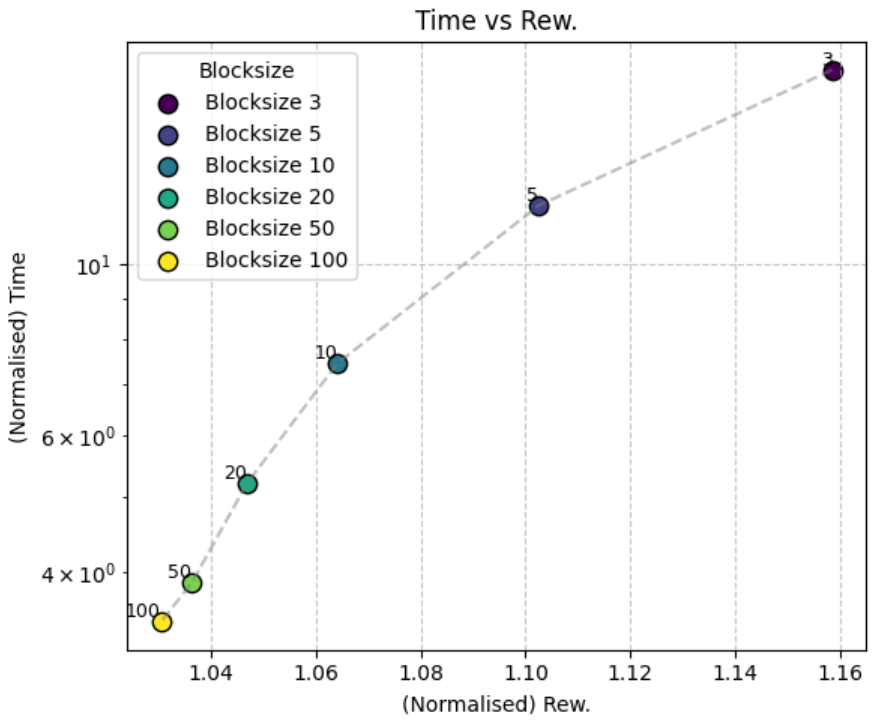}
    \caption{Blocksize Analysis for Time vs Reward}
      \label{fig:same_blocksize_1}
  \end{subfigure}
  \hfill
  \begin{subfigure}[b]{0.48\linewidth}
    \includegraphics[trim=100 250 100 250 ,width=1.0\linewidth]{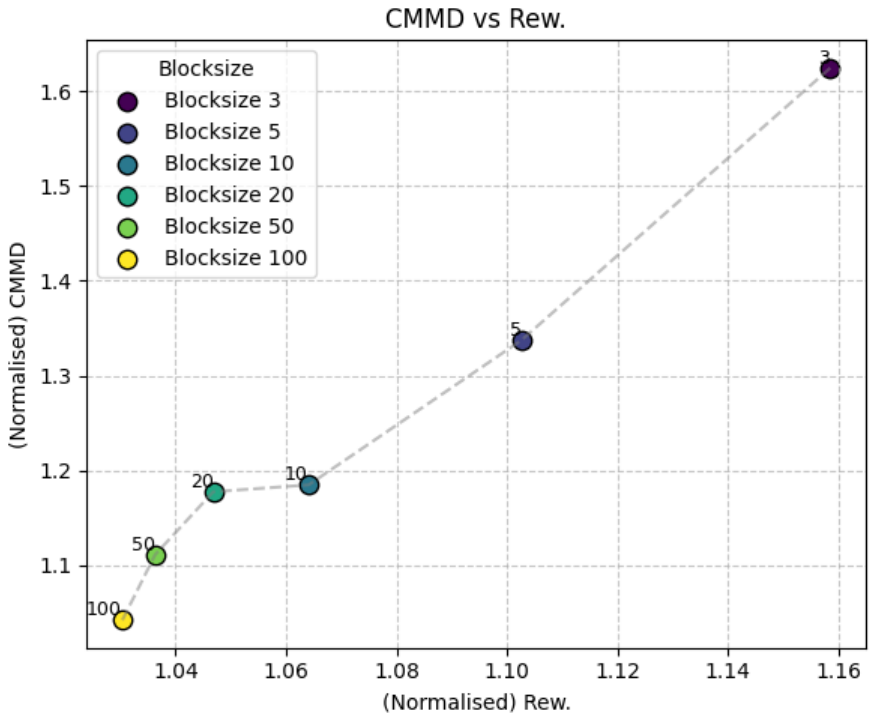}
    \caption{Blocksize Analysis for Divergence vs Reward}
  \label{fig:same_blocksize_2}
  \end{subfigure}
\end{figure}

In this experiment, we set both blocksizes equal and analyze the resulting trade-offs, such as Reward vs. Divergence and Reward vs. Time (see Figures~\ref{fig:same_blocksize_1}, \ref{fig:same_blocksize_2}). Decreasing the sampling blocksize ($B_s$) leads to more aggressive sampling, as selection occurs more frequently, while simultaneously aligning the prior more closely with the posterior by incorporating gradients at finer intervals (decreasing the gradinet blocksize $B_g$). This improves reward alignment, but comes at the cost of higher divergence and longer runtime.

\subsection{Constant $B_g$, Varying $B_s$}
For this experiment, we keep the gradient block size equal ($B_g=5$) and analyze how changing the sampling blocksize affects the performance through the resulting trade-offs, such as Reward vs. Divergence and Reward vs. Time (see Figures~\ref{fig:same_blocksize_1},\ref{fig:same_blocksize_2}). As expected, decreasing the blocksize leads to more frequent selection and replication of the best sample across all streams. This results in increased exploitation, which manifests as higher rewards, at the cost of higher divergence and increased computational time.

\begin{figure}[ht]
  \centering
  \begin{subfigure}[b]{0.48\linewidth}
    \includegraphics[trim=100 250 100 250 ,width=1.0\linewidth]{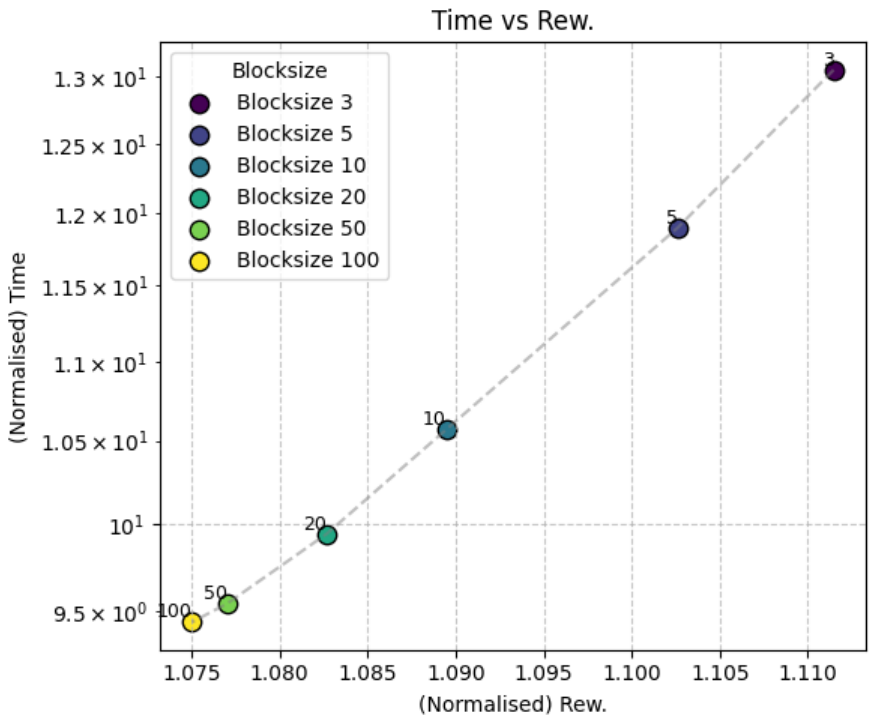}
    \caption{Blocksize Analysis for Divergence vs Reward}
      \label{fig:same_blocksize_bg1}
  \end{subfigure}
  \hfill
  \begin{subfigure}[b]{0.48\linewidth}
    \includegraphics[trim=100 250 100 250 ,width=1.0\linewidth]{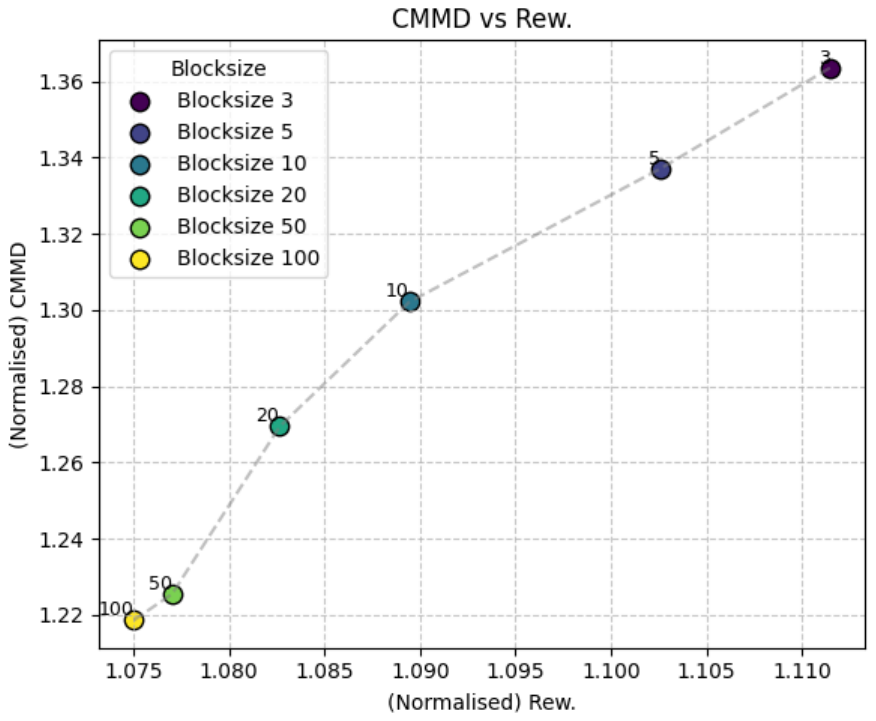}
    \caption{Blocksize Analysis for Time vs Reward}
  \label{fig:same_blocksize_bg2}
  \end{subfigure}
\end{figure}

\subsection{Constant $B_s$, Varying $B_g$}
For this experiment, we keep the gradient block size equal ($B_s=5$) and analyze how changing the sampling blocksize affects the performance through the resulting trade-offs, such as Reward vs. Divergence and Reward vs. Time (see Figures~\ref{fig:same_blocksize_1},\ref{fig:same_blocksize_2}). As the blocksize decreases, gradients are incorporated at more timesteps, which aligns the prior more closely with the posterior. This results in improved rewards, but comes at the cost of increased divergence and longer computation time.

\begin{figure}[ht]
  \centering
  \begin{subfigure}[b]{0.48\linewidth}
    \includegraphics[trim=100 250 100 250 ,width=1.0\linewidth]{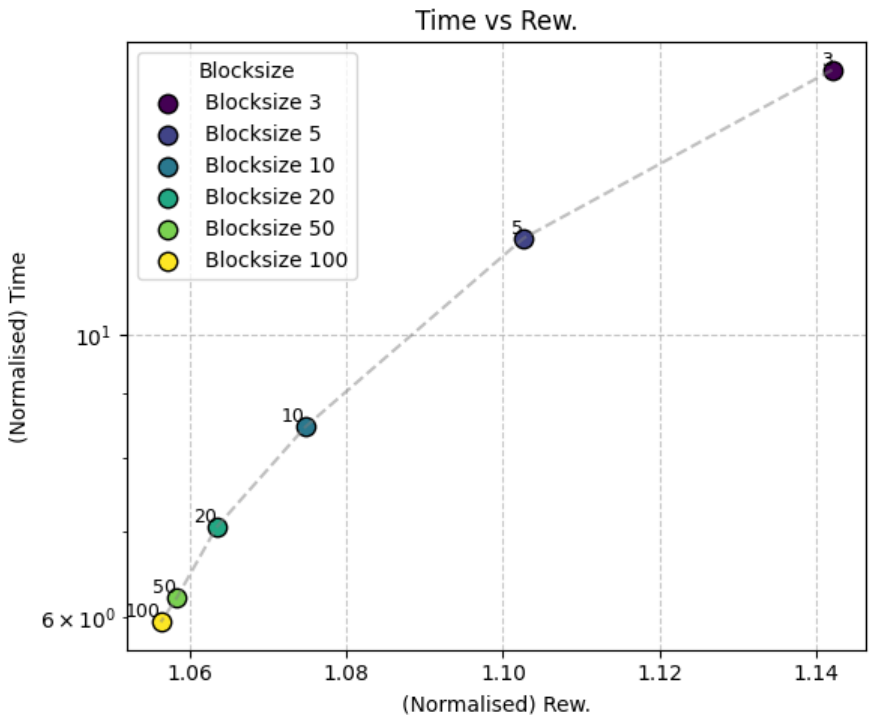}
    \caption{Blocksize Analysis for Divergence vs Reward}
      \label{fig:same_blocksize_bs1}
  \end{subfigure}
  \hfill
  \begin{subfigure}[b]{0.48\linewidth}
    \includegraphics[trim=100 250 100 250 ,width=1.0\linewidth]{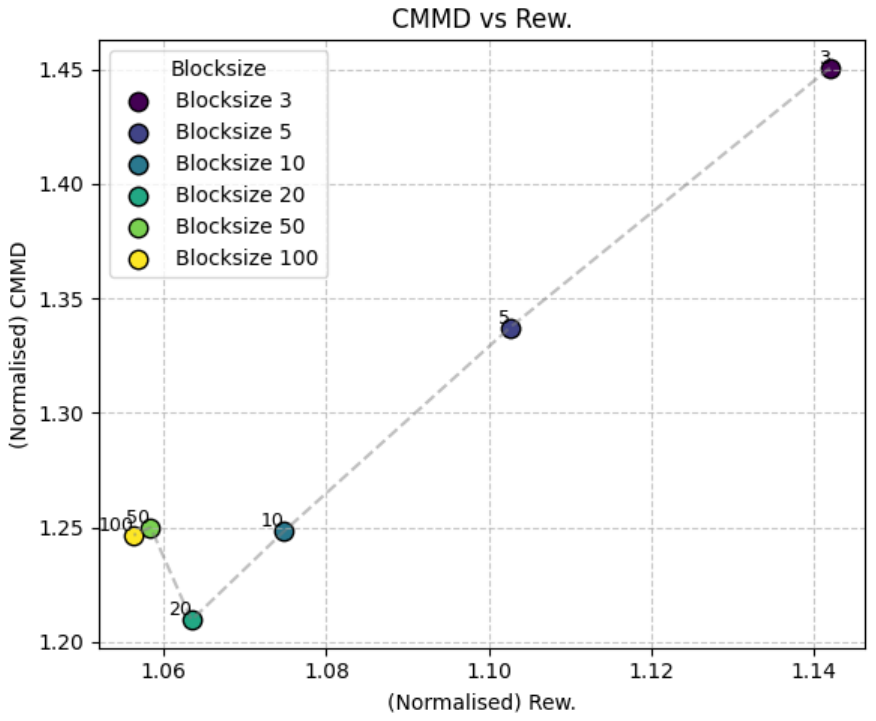}
    \caption{Blocksize Analysis for Time vs Reward}
  \label{fig:same_blocksize_bs2}
  \end{subfigure}
\label{}
\end{figure}

\section{General Guidelines for Setting $N$, $B_g$, $B_s$}
\texttt{UniCoDe} depends heavily on the parameters $N$, $B_g$, and $B_s$, which intuitively control the degree of exploration and the extent to which the prior is pushed towards the posterior (via $B_g$). As with $N$ and $B$ in \texttt{CoDe} \cite{singh2025code} Appendix Section G, these parameters influence both reward-alignment and fidelity to the base distribution, as illustrated in Figs.~1-4. 

Increasing $N$ enhances exploration, leading to higher reward-aligned generations, but also increases divergence from the prior distribution and computational cost (Figs.~\ref{fig:rew_vs_div} and \ref{fig:rew_vs_time}). Similarly, decreasing $B_g$ significantly boosts rewards, as observed in Figs.~\ref{fig:same_blocksize_bs1} and \ref{fig:same_blocksize_bs2}, but this comes at the cost of reward hacking, reflected in higher divergence from the prior. $B_s$ affects the frequency of sampling, with lower values increasing reward-alignment by being more aggressive at the cost of raising divergence and computational requirements (Figs.~\ref{fig:same_blocksize_bg1} and \ref{fig:same_blocksize_bg2}). 

Overall, the interplay between $N$, $B_g$, and $B_s$ governs the trade-off between reward maximization, adherence to the base distribution and the computational time.

\section{Additional Results}

\begin{figure}[H]
  \centering
  \includegraphics[trim=0 120 0 120,clip,width=1.0\linewidth]{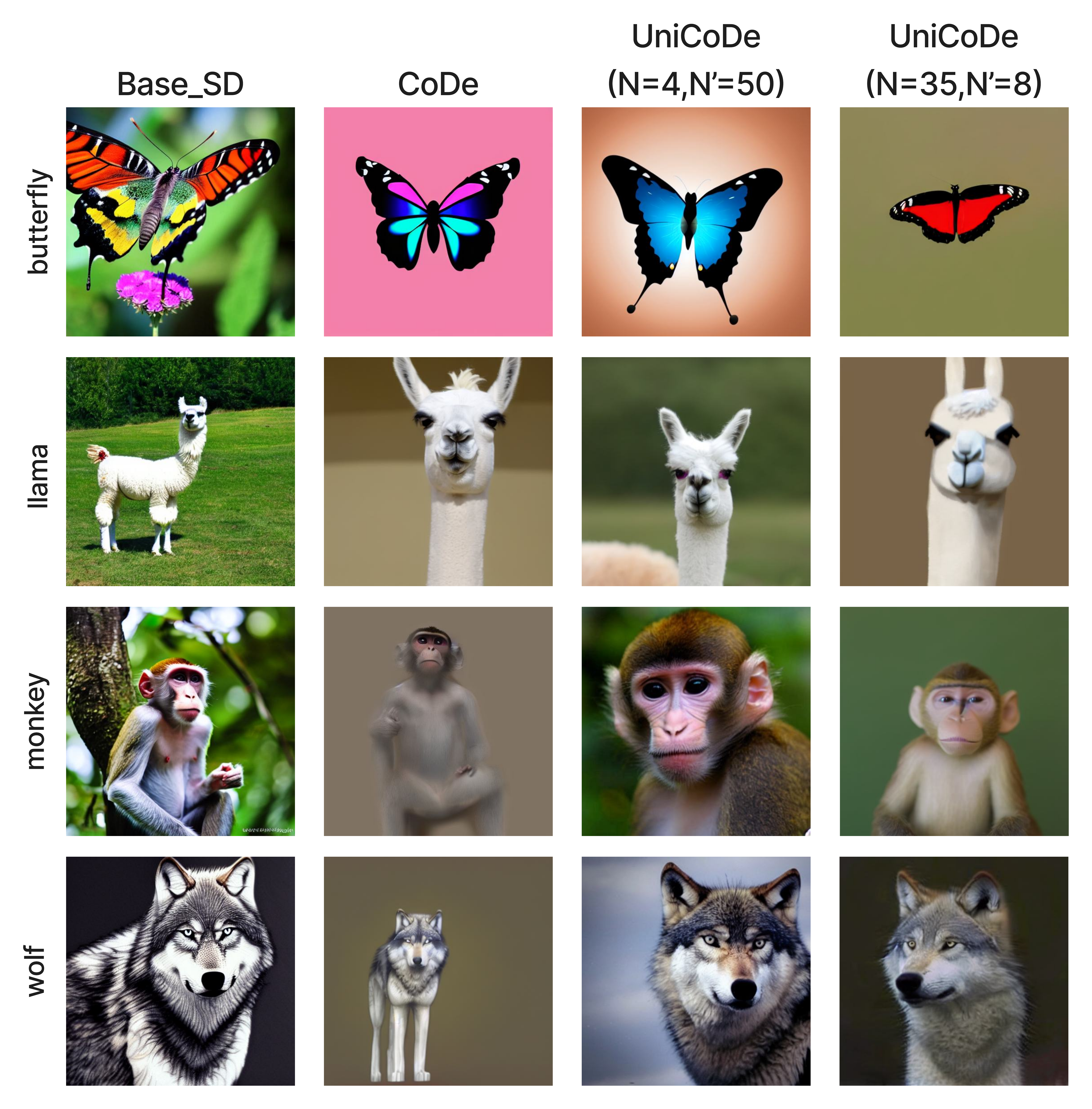}
  \caption{More qualitative samples for the compressibility scenario (non-differentiable)}
\label{fig:ablation-comp}
\end{figure}

\begin{figure}
  \centering
  \includegraphics[trim=0 70 0 70,clip,width=1.0\linewidth]{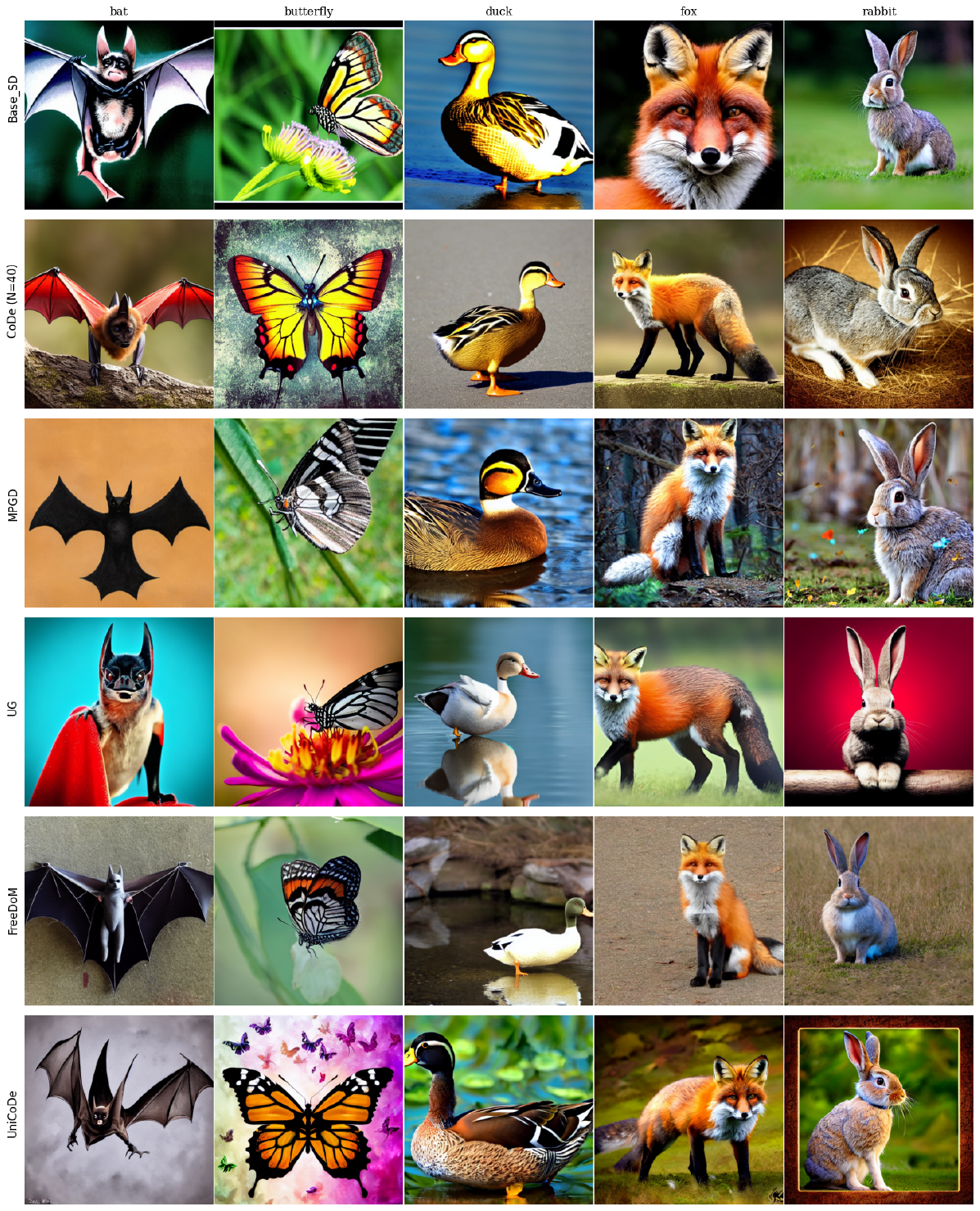}
  \caption{More qualitative examples for aesthetic guidance}
\label{fig:qualitative-aesthetic-final}
\end{figure}

\begin{figure}
  \centering
  \includegraphics[trim=10 70 10 70,clip,width=1.0\linewidth]{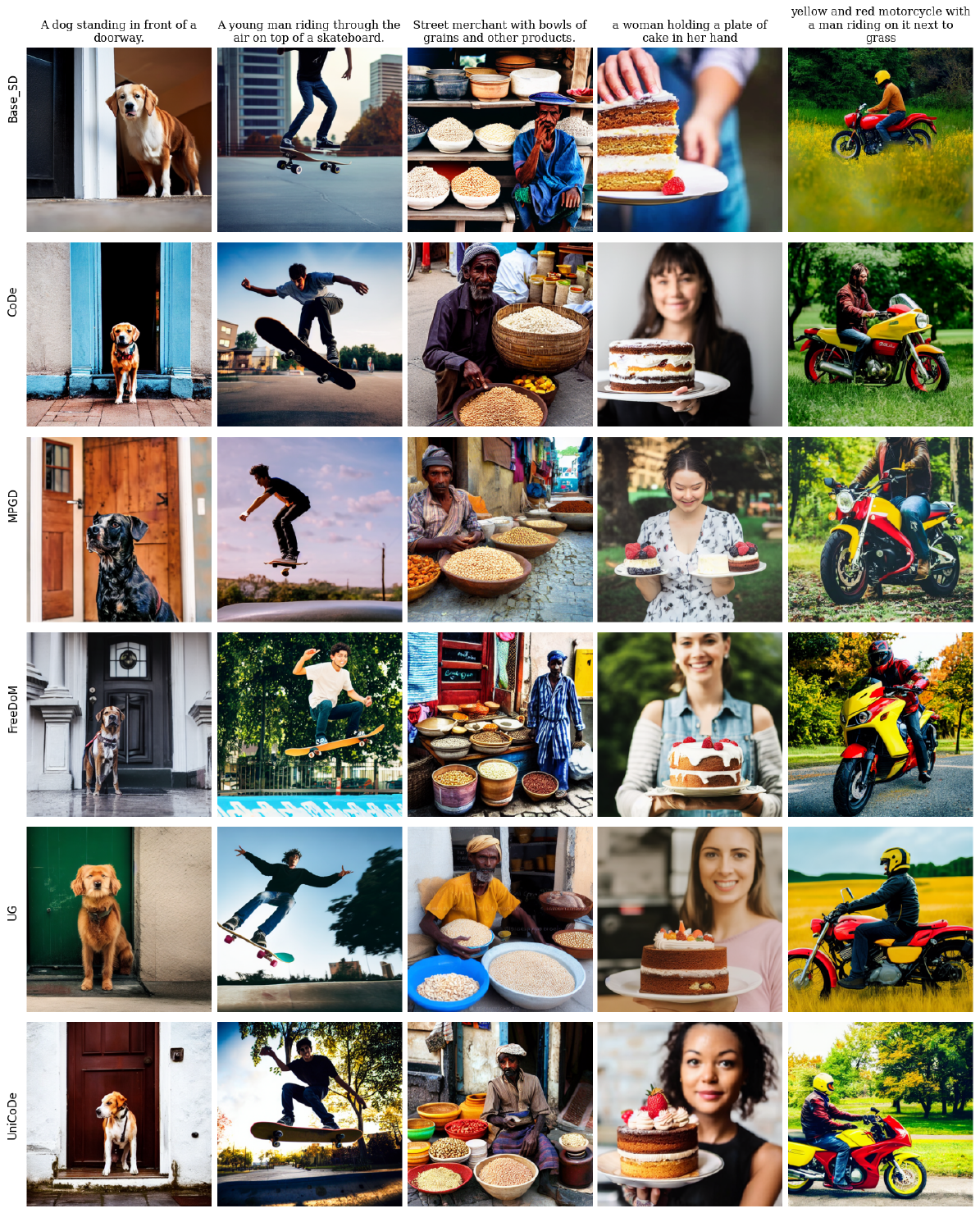}
  \caption{More qualitative examples for T2I pickscore guidance}
\label{fig:qualitative-pic-final}
\end{figure}

\begin{figure}[t]
  \centering
  \includegraphics[trim=0 20 0 20,clip,height=0.95\textheight]{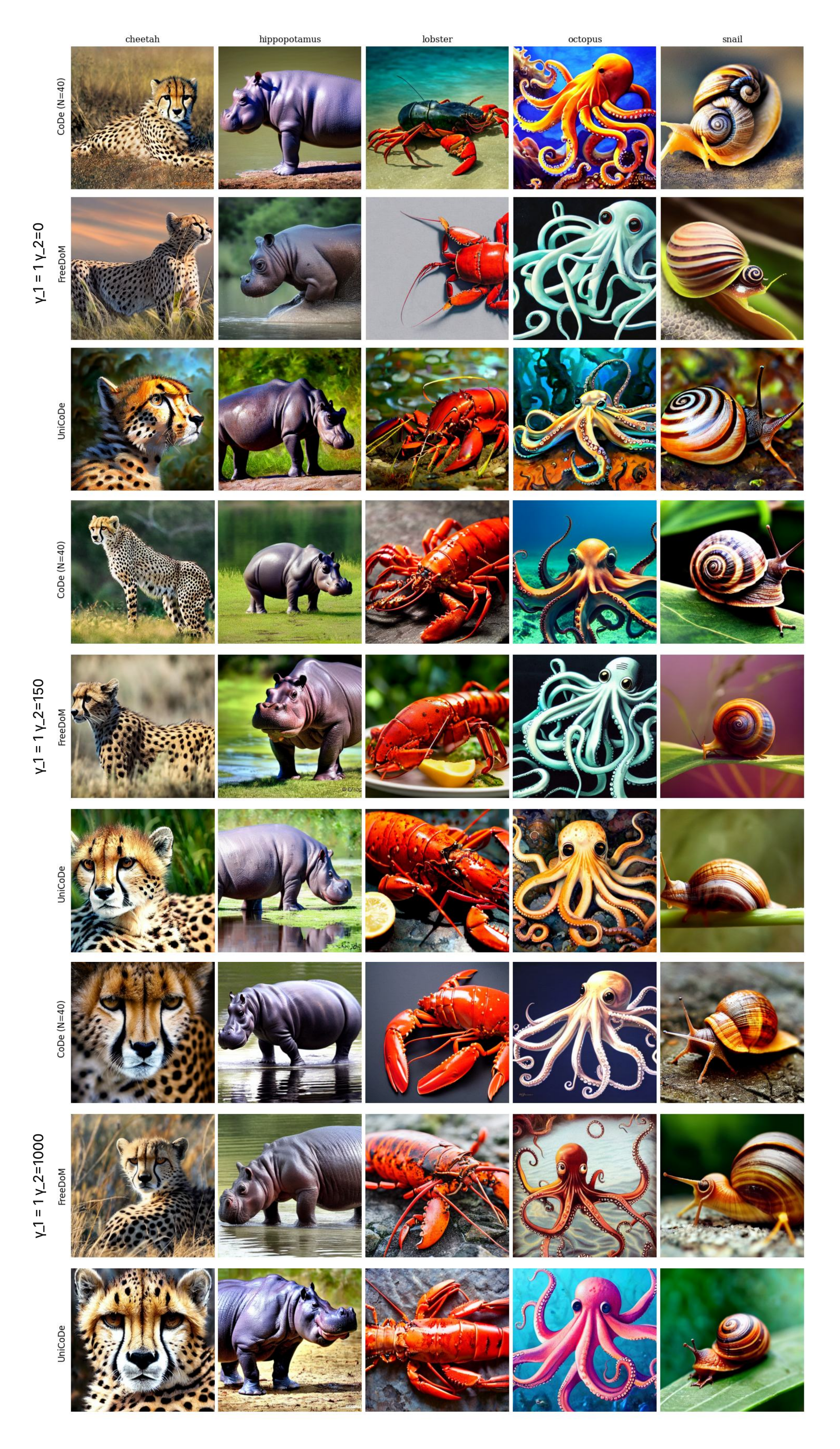}
  \caption{Each column corresponds to a different prompt. Rows are grouped by $\gamma_2 \in \{0, 150, 1000\}$ (top to bottom), with $\gamma_1$ fixed at $1$. Within each group, the three rows show results from \texttt{CoDe}, \texttt{FreeDoM}, and \texttt{UniCoDe} (top to bottom). The top group focuses more on aesthetic quality, whereas the bottom group prefers Pickscore, and the middle group ($\gamma_2 = 150$) offers a balanced trade-off.}
  \label{fig:ablation-multi-final}
\end{figure}

\begin{figure}
  \centering
  \includegraphics[trim=0 100 0 120,clip,width=0.95\linewidth]{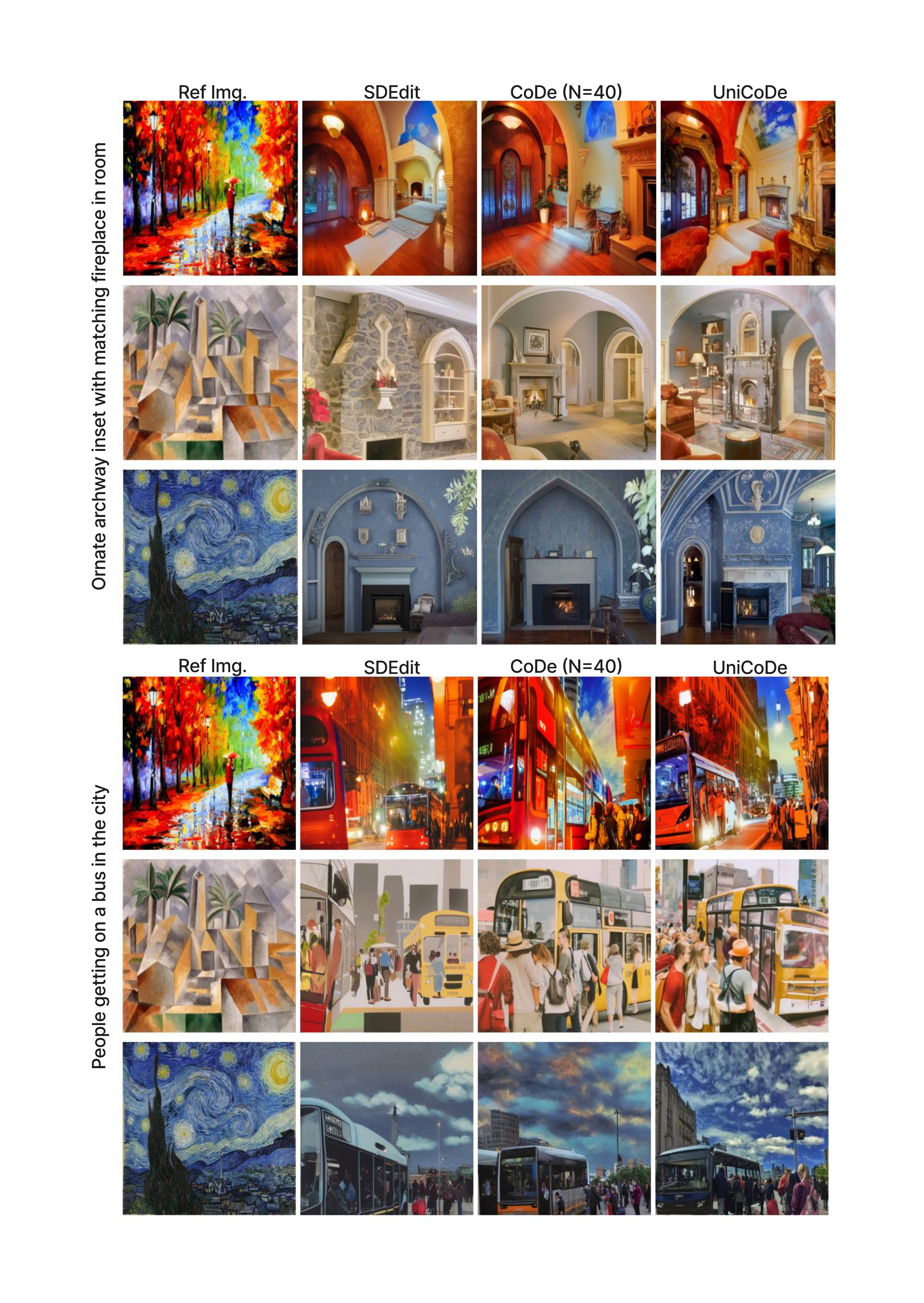}
  \caption{More qualitative samples for the (T+I)2I  scenario}
\label{fig:ablation-it2i}
\end{figure}




\clearpage

{
    \small
    \bibliographystyle{unsrt}
    \bibliography{main}
}

\end{document}